\documentclass[review]{elsarticle}
\usepackage{lineno,hyperref}
\modulolinenumbers[5]

\journal{Elsevier}

\usepackage{url}
\usepackage{makeidx}         % allows index generation
\usepackage{graphicx}        % standard LaTeX graphics tool
\usepackage{multicol}        % used for the two-column index
\usepackage[bottom]{footmisc}% places footnotes at page bottom

\usepackage{amsmath,bm,amsfonts,amssymb}

\usepackage{commath}      % for norm 
\usepackage{color,xcolor,ucs}
\usepackage{subfig}
\usepackage[font=small,labelfont=bf]{caption}

\usepackage{floatrow}
\usepackage{tabularx}
\usepackage{float}
\usepackage{xr-hyper}
\usepackage{wrapfig}

\usepackage{algorithm}
\usepackage{algpseudocode}% http://ctan.org/pkg/algorithmicx
\usepackage{multirow}
\usepackage{listings}
\usepackage{lipsum}
\usepackage{textcomp} % for registered symbol

\usepackage[nocomma]{optidef}  % for optimization objective function definition
\newtheorem{definition}{Definition}

\definecolor{revisioncolor}{rgb}{0.1,0.1,1}

\begin{document}

\begin{frontmatter}
%
% \title{Locally Optimal Solutions to Constraint Displacement Problems via Constraint Overlaps}
%
\title{Locally Optimal Solutions to Constraint Displacement Problems via Path-Obstacle Overlaps}

\author[label1]{Antony Thomas}
\ead{antony.thomas@iiit.ac.in}
\address[label1]{Robotics Research Center, IIIT Hyderabad, Hyderabad 500032, India.}

\author[label2]{Fulvio Mastrogiovanni}
\ead{fulvio.mastrogiovanni@unige.it}
\address[label2]{Department of Informatics, Bioengineering, Robotics, and Systems Engineering, University of Genoa, Via All'Opera Pia 13, 16145 Genoa, Italy. }

\author[label2]{Marco Baglietto}
\ead{marco.baglietto@unige.it}

\begin{abstract}
We present a unified approach for constraint displacement problems in which a robot finds a feasible path by displacing constraints or obstacles. To this end, we propose a two stage process that returns locally optimal obstacle displacements to enable a feasible path for the robot. The first stage proceeds by computing a trajectory through the obstacles while minimizing an appropriate objective function. In the second stage, these obstacles are displaced to make the computed robot trajectory feasible, that is, collision-free. Several examples are provided that successfully demonstrate our approach on two distinct classes of constraint displacement problems. 
\end{abstract}

\begin{keyword}
Constraint Displacement Motion Planning, Constraint Overlaps, Minimum Constraint Displacement, Minimum Constraint Removal.
\end{keyword}
\end{frontmatter}
\section{Introduction}
As humans, we encounter various situations in our day to day life in which we alter the location of objects -- opening closed doors, repositioning chairs or other movable objects, clear objects while picking an object of interest from a cluttered table-top. As opposed to avoiding each object, altering or displacing these objects or constraints allow us to expand the solution space of feasible paths. Such an expansion of the solution space is called for in several robot applications, for example, when the classical obstacle avoidance fails due to non existence of a feasible path. In such situations, constraints, such as movable obstacles, may be cleared to find feasible paths. Manipulators often need to rearrange or move obstacles aside to accomplish a given set of tasks -- a futuristic robot cooking dinner at home, manipulation in industrial settings, shelves replenishment in a grocery store. Service robots may need to reposition chairs or other movable objects to accomplish a task. A robot may need to plan a path through dynamic obstacles as they might clear the path while moving.

We define a \textit{constraint displacement problem} as one that finds a feasible path by displacing constraints while minimizing a problem-specific objective function. While different problems exist, which are characterized by analogous objectives, a few of these are discussed below.
\begin{itemize}
\item The minimum constraint displacement (MCD)~\cite{hauser2013RSS} problem aims to find a feasible path while minimizing obstacle displacement magnitudes.
\item The minimum constraint removal (MCR)~\cite{hauser2014IJRR} problem seeks to determine the minimum number of movable obstacles to be removed (or displaced) to obtain a feasible path.
\item Navigation among movable obstacles (NAMO)~\cite{stilman2005IJHR} focuses on finding a feasible path while minimizing the number of moved obstacles and the work done by the robot during the rearrangement of these obstacles to clear a path. We note here that NAMO problems aim to find sequences of paths for robot motion and object manipulation, and they can have any general objective function. 
\item In scenarios where a target object needs to be retrieved from clutter, a set of obstacles is rearranged through manipulation under clutter~\cite{dogar2011RSS}.
\end{itemize}  

In this paper, we present an integrated approach for constraint displacement problems that proceeds in two steps.
In the first step, the robot plans a path through the movable obstacles while minimizing the problem specific objective function. We thus allow for \textit{constraint overlaps}, with the extent and the number of such overlaps decided by the objective function. For example, in MCD the overall overlap with the obstacles is to be minimized as greater overlap implies greater constraint displacements. In the case of MCR and manipulation in clutter, the objective is to formulate it in a way that minimizes the total number of unique constraint overlaps. Note that the extent of overlap is a measure of the desired distance an obstacle is to be moved. The distance to be moved in combination with the force to be applied at a contact point may be used to compute the work done by the robot. Adding the amount of work to the MCR cost gives a variant of the NAMO problem.
In the second step, overlapping obstacles are displaced so that the planned trajectory is collision-free. 
To this end, we compute locally optimal constraint displacements encoding both obstacles' translation and rotation. 

\begin{figure}[t]
  \subfloat[]{\includegraphics[trim=52 60 38 60,clip,scale=0.5]{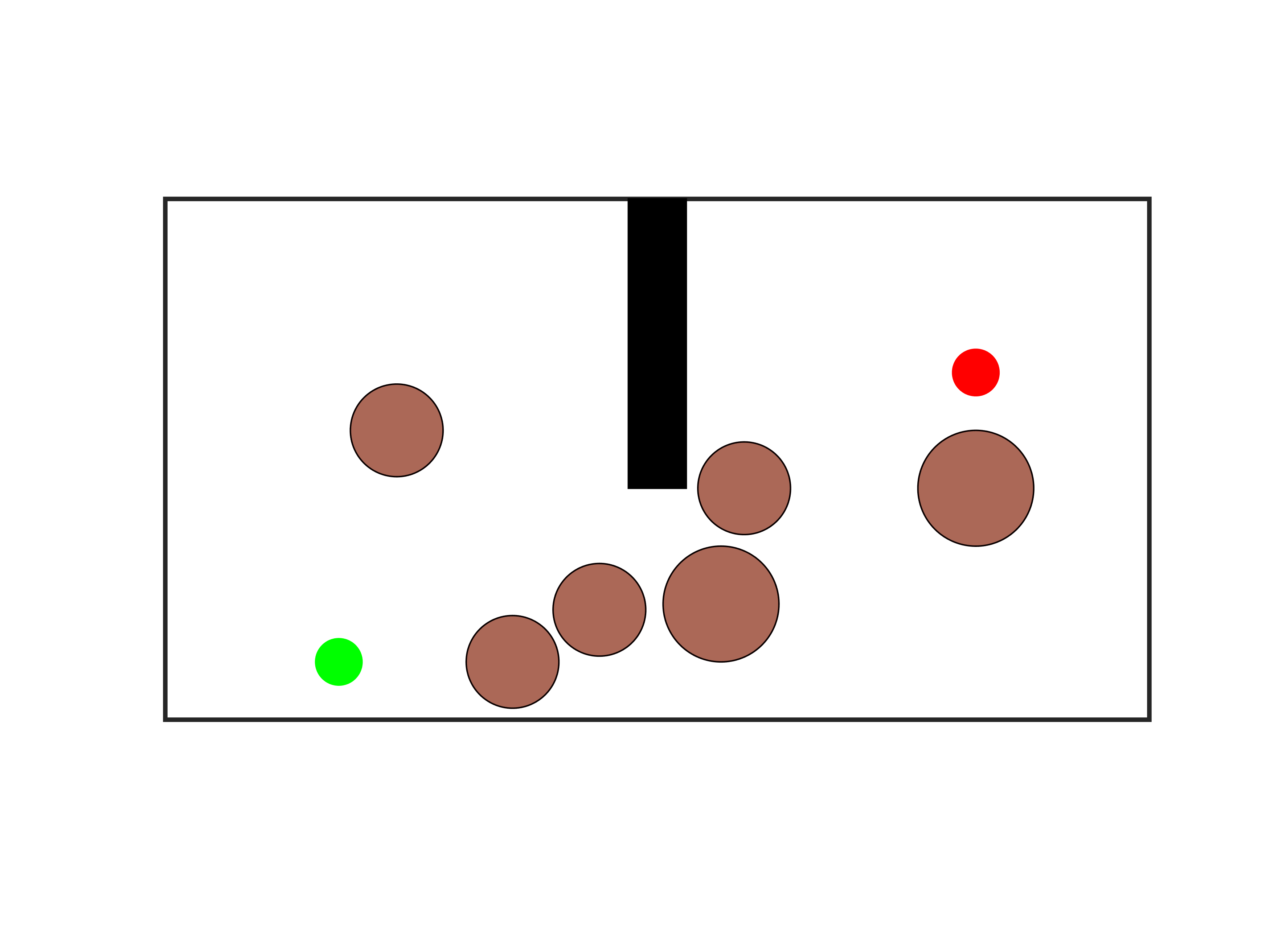}\label{fig:ras1}}\hfill
   \subfloat[]{\includegraphics[trim=52 57 38 60,clip,scale=0.49]{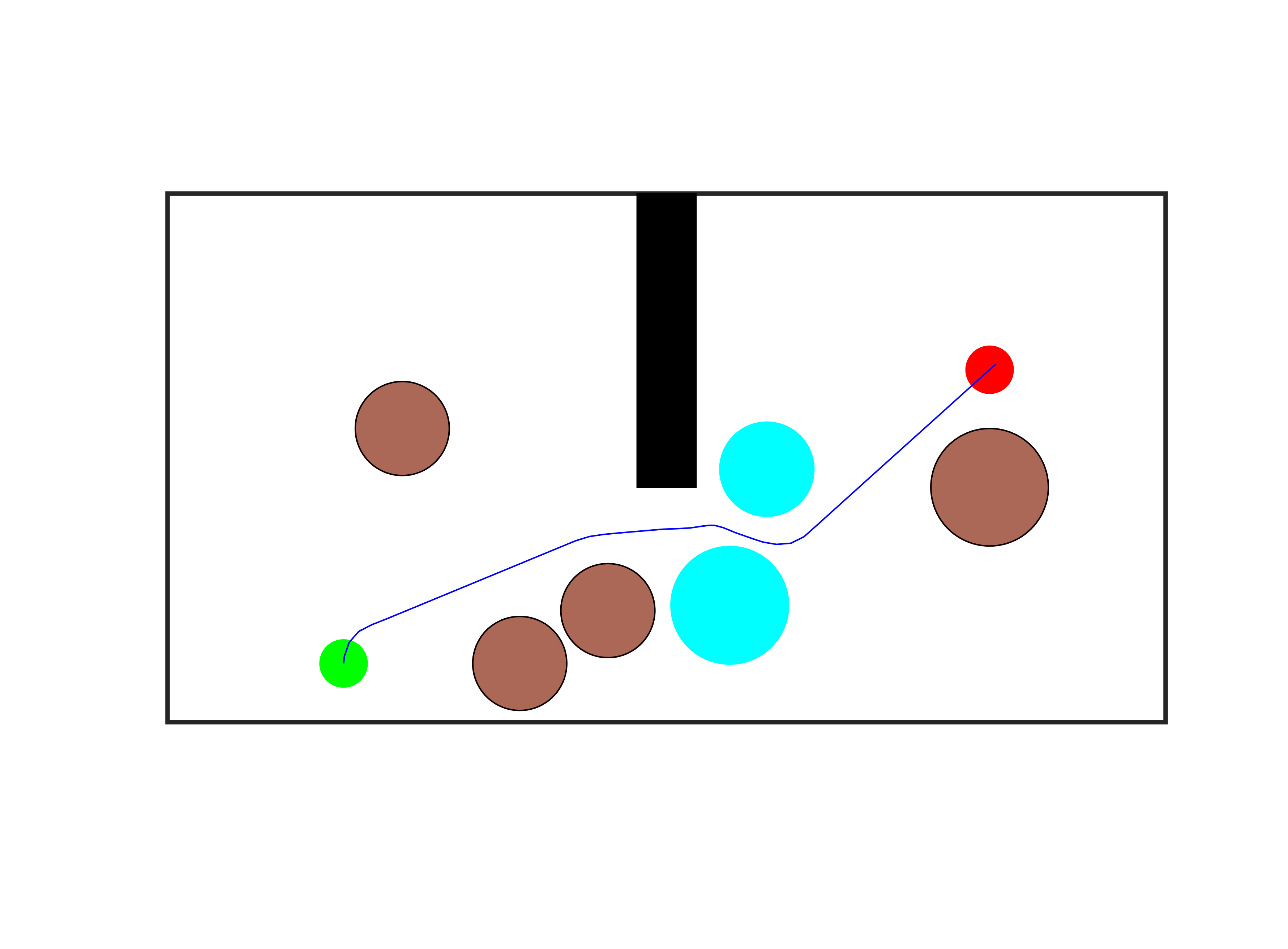}\label{fig:ras2}}\hfill
  \subfloat[]{\includegraphics[trim=52 60 38 60,clip,scale=0.5]{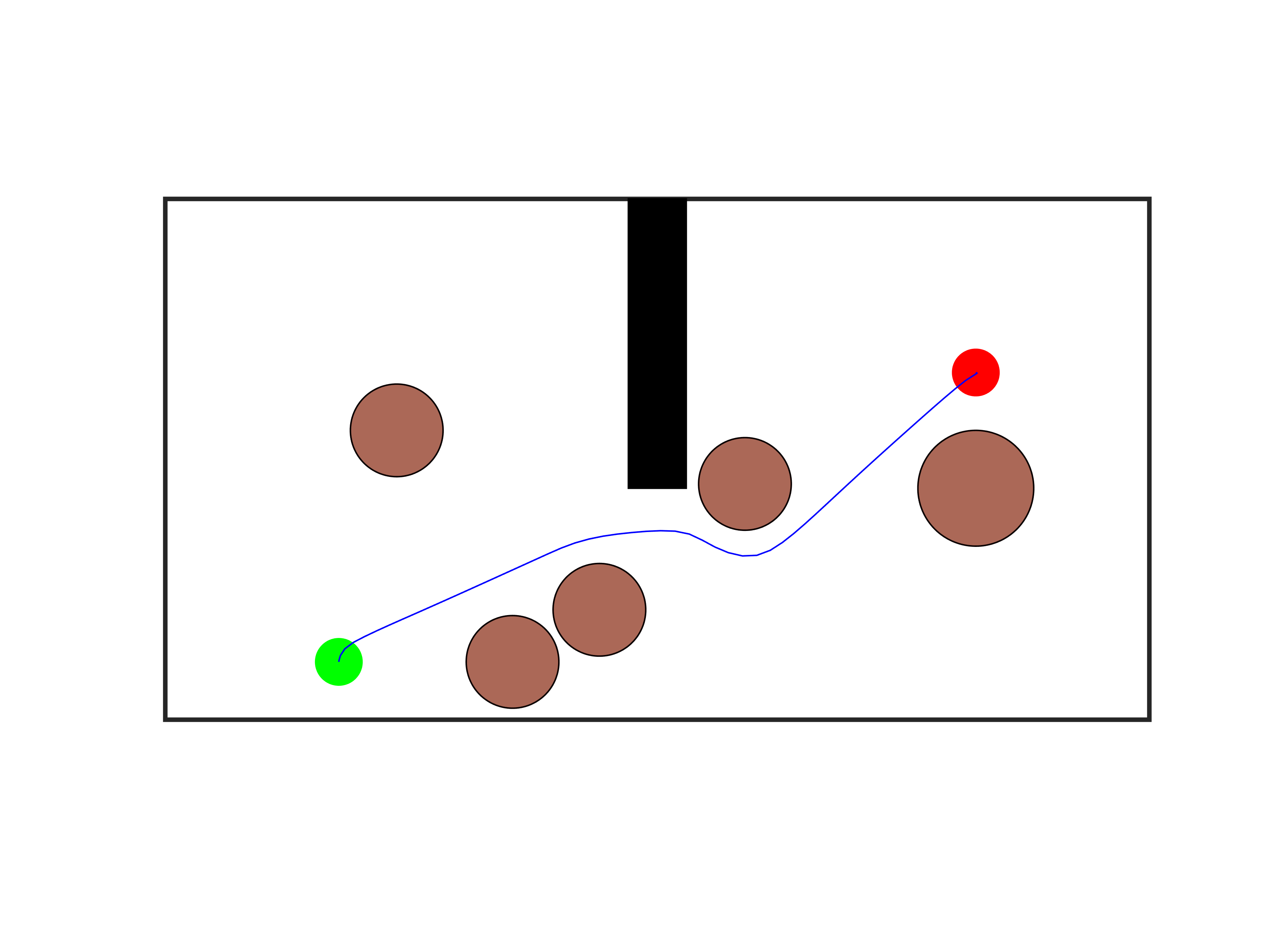}\label{fig:ras3}}\hfill
  \subfloat[]{\includegraphics[trim=52 60 38 60,clip,scale=0.5]{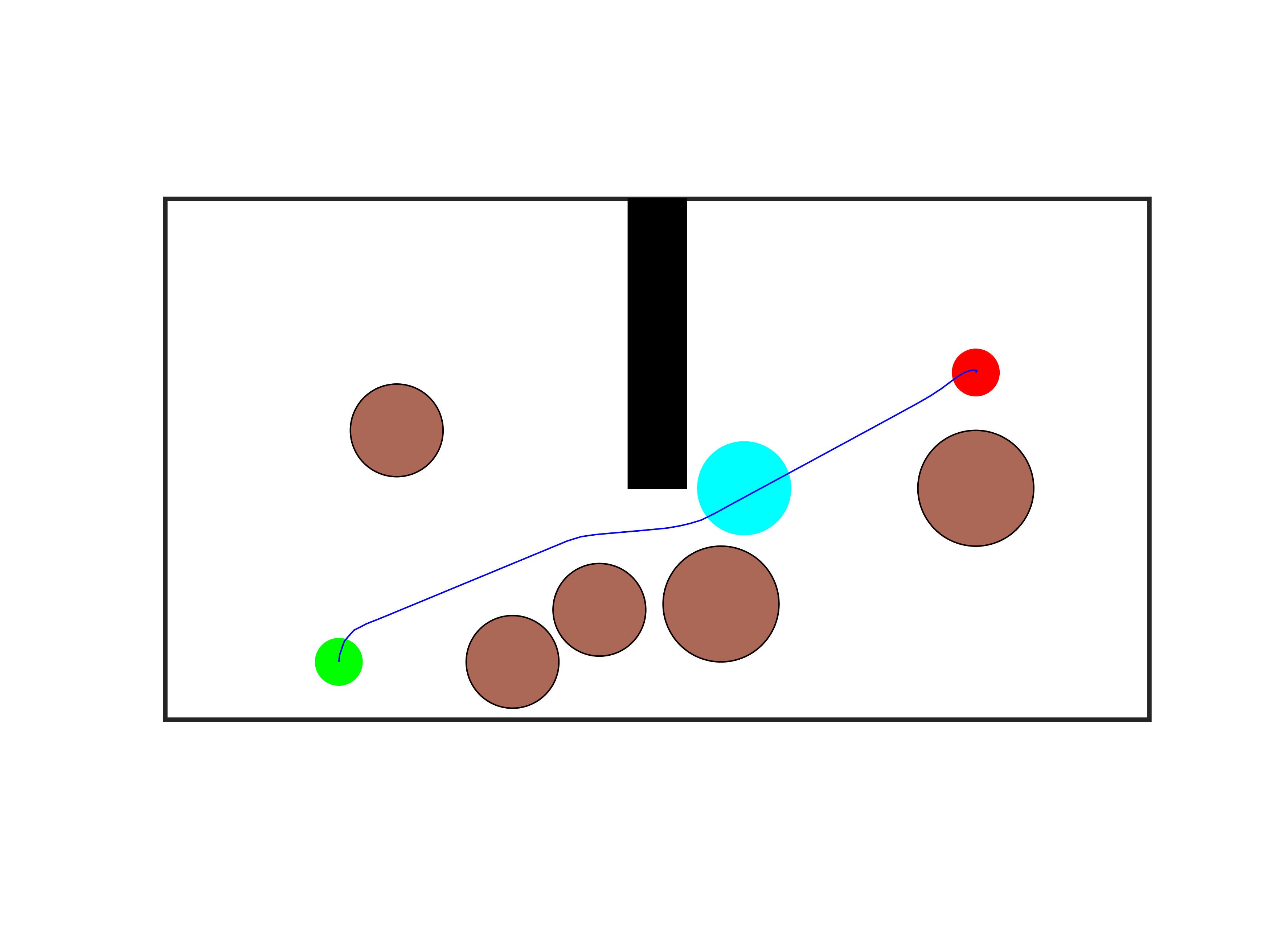}\label{fig:ras4}}\hfill
     \caption{(a) The robot at the starting point (green) needs to move to the goal (red) while navigating through movable obstacles. (b) The MCD path is depicted with displaced obstacles shown in cyan. (c) One obstacle has been removed from the environment to create a collision-free path, without the need to displace any obstacles. Our approach solves the classical motion planning problem and computes a path without moving any obstacles. (d) The cost function has been modified to address the MCR problem. The cyan object needs to be removed for a feasible path to exist.}
  \label{fig:ras}
\end{figure}

Using the approach described above, potential solutions to various classes of constraint displacement problems are obtained by employing apposite objective functions (Fig.~\ref{fig:ras}). This paper specifically focuses on MCD and MCR problems in the context of mobile robot navigation. Hauser~\cite{hauser2014IJRR} demonstrated the NP-hardness of MCR via a reduction from the SET-COVER problem. In this formulation, the MCR problem in the robot’s configuration space is transformed into a graph-based representation. Intuitively, such a graph problem can be addressed using a best-first search strategy. However, when obstacles overlap in configuration space, paths may enter or exit an obstacle multiple times, and in the worst case, the best-first search may generate a number of states exponential in the number of obstacles. The NP-hardness of MCD follows directly through a trivial reduction from MCR~\cite{hauser2013RSS}. Hence, it is appropriate to consider approximate or heuristic solution strategies for these problems.

Experimental results for variants of MCD and MCR problems are presented by appropriately modifying their objective functions. In the present work, we do not focus on the specific mechanics of obstacle displacement (that is, we ignore factors such as mass, shape, or required interaction forces) and assume that the computed displacements can be achieved irrespective of the type of robot-obstacle interaction. In terms of theoretical development, this allows us to study a 2D domain using planar projections of the 3D environment.

An initial version of this work appeared in~\cite{thomas2022IAS}. This work is extended in the following manner.
\begin{itemize}
\item Generalization to constraint displacement problems, as opposed to the MCD formulation in~\cite{thomas2022IAS}.
\item Efficient displacement of overlapping obstacles by solving an optimization problem, as opposed to the iterative displacement from a discrete displacement set in~\cite{thomas2022IAS}.
\item Additional experiments demonstrating the applicability of our approach to MCD and MCR problems.
\end{itemize}

The paper is organized as follows.
Section \ref{sec:related_work} introduces the relevant background.
A more formal definition of the problem we investigate is discussed in Section \ref{sec:problem_definition}.
Section \ref{sec:approach} presents and discusses our approach, whereas a possible implementation is introduced in Section \ref{sec:implementation}. 
A validation is reported in Section \ref{sec:results}.
Conclusions follow.

\section{Related Work}
\label{sec:related_work}
Constraint displacement refers to a broad class of motion planning problems. If no displacements are allowed then the problem reduces to the classical motion planning problem or the Piano Mover's problem~\cite{reif1979CS}. MCD is an NP-hard problem that examines how a robot can reach its goal by displacing obstacles while minimizing the overall displacement of obstacles. Hauser~\cite{hauser2013RSS} samples both robot configurations and obstacle displacements and iteratively increments the samples until the best cost solution is obtained. The MCD approach in~\cite{thomas2023ICRA} plans a trajectory that considers robot dynamics and other constraints, including control limits and obstacle avoidance. It aims to find the sequence of control actions that move the robot to the goal while minimizing the displacement of obstacles. Introduced by Hauser in~\cite{hauser2014IJRR}, MCR seeks to find the minimum number of obstacles or constraints to be removed to obtain a feasible path. 
The approach in~\cite{hauser2014IJRR} employs a sampling-based approach to sample robot configurations and incrementally grows the roadmap. The path in the roadmap passing through a minimum number of obstacles is the required solution to the MCR problem. 
However, MCR does not indicate where the obstacles should be moved. 
The MCR problem has received considerable attention in the past and different variants of the problem exist in the literature~\cite{castro2013CDC,krontiris2017AR,eiben2018AAAI,kumar2019AEA,xu2020ICMLC,thomas2023IAS}. 
This problem is shown to be NP-hard for navigation in the plane even when the obstacles are all convex polygons~\cite{erickson2013AAAI}. 

Similar challenges are exhibited in the context of task and motion planning~\cite{kaelbling2013IJRR,srivastava2014ICRA,dantam2016RSS,garrett2018IJRR,thomas2021RAS}. 
To pick an object from a shelf, a robot may need to consider displacing different \textit{constraints} if the object is at the far end of the shelf. 
Selecting the obstacles to be displaced may be accomplished via task planning and the geometric constraints associated with their displacements. 
Another similar class of problems is manipulation among clutter or rearrangement planning in clutter~\cite{stilman2007ICRA,dogar2011RSS,krontiris2015RSS,karami2021AIIA}. 
Constraints may hinder the end-effector’s reachability workspace, and therefore they need to be displaced to pick a target of interest.
Among these, the work in~\cite{dogar2011RSS} is more closely related since~\cite{dogar2011RSS} first finds an action that moves the target object to the goal configuration. 
The total volume of space the robot and the manipulated object need during the action is then computed and the overlapping obstacles are moved aside. 
NAMO~\cite{stilman2005IJHR,nieuwenhuisen2008WAFR} is a related class of problems and is proved to be NP-hard~\cite{wilfong1988ASCS}. 
Most approaches solve a subclass of problems, selecting a set of obstacles to be moved to reconfigure the environment.
~\cite{stilman2005IJHR} consider additional aspects such as finding contact points for grasping or pushing the object to be moved. 
Proving path non-existence in motion planning~\cite{zhang2008WAFR,basch2001ICRA,li2021RSS} is another related area since these approaches try to find constraints preventing robot motions from start to goal.

In this work, we present a unified framework that can model various constraint displacement problems discussed above by employing the appropriate objective functions. 
Specifically, we focus on finding approximate solutions to MCD and MCR problems and demonstrate the applicability of our approach to this class of problems.

\section{Problem Definition}
\label{sec:problem_definition}
Let $\mathcal{C}$ denote the robot configuration space with $x^s, x^g \in \mathcal{C}$ denoting the start and goal states. 
Let $\mathcal{O}= \{o^i| \ 1 \leq i \leq m\}$ denote the set of obstacles in the environment with $n<=m$ obstacles that are movable. 
By abuse of notation we will use $o^i$ to denote both the $i$-th obstacle as well as its state. 
The obstacles are associated with a displacement set $\mathcal{D}= \{d^i| \ 1 \leq i \leq n\}$, which represents the corresponding obstacle displacements. 
The new obstacle location after being displaced by $d^i$ will be denoted by $o^i(d^i)$. 
In this paper, we will consider obstacles that can either rotate or translate or perform both, and therefore $d^i$ belongs to a displacement space of arbitrary dimension. 
We also define an overlap set $\mathcal{V}= \{v^i| \ 1 \leq i \leq n\}$ that denotes the overlap of each movable obstacle with the robot.
\begin{definition}
Let us consider a robot that must navigate from $x^s$ to $x^g$ but is unable to find a feasible path due to constraints in the environment. In situations in which the path is blocked by movable obstacles, a constraint displacement motion planning problem from $x^s$ to $x^g$ finds a path $x_{0:T}= \{x_0,\ldots,x_T\}$ in $\mathcal{C}$ ($x_0 = x^s$) with obstacles displacements $d^1,\ldots,d^n$.
\label{def:cd}
\end{definition}
This paper investigates a possible solution to constraint displacement problems with an approach that proceeds in two stages, namely the overlap stage and the displacement stage.
\begin{definition}
\textbf{Overlap stage.}
The overlap stage plans a trajectory through the movable obstacles such that the robot minimizes a weighted sum of robot state cost, control cost and overlap costs given by
\label{def:overlap}
\end{definition}
\begin{equation}
J = w^xc(x_{0:T}) + w^v \sum_{i=1}^n c^i(v^i) + w^uc(u_{0:T-1})
\label{eq:overlap_stage_cost}
\end{equation}
\noindent with $c(x_{0:T})$ representing a general robot state-dependent cost, $c^i(v^i)$ denoting the overlap cost of the $i$-th overlapping obstacle, $c(u_{0:T-1})$ accounting for the cost of the control action or actions executed by the robot, whereas $w^x$, $w^v$, and $w^u$ serving as the corresponding weights.
The summation is to be performed over all the movable obstacles. 
The overlap cost is a conditional cost given by
\begin{equation}
c^i(v^i) = 
   \begin{cases}
     c^i(v^i) \ &\text{if overlap, that is,} \ v^i \neq 0 \\
     0 \ &\text{if no overlap, that is,} \ v^i = 0.
   \end{cases}
\label{eq:overlap_cost}
\end{equation}
\begin{definition}
\textbf{Displacement stage.} 
For each $v^i \neq 0$, this stage finds locally optimal displacements $d^i$ such that  
\label{def:displacement}
\begin{equation}
x_t \cap \bigcup_{i=1}^n o^i(d^i) = \{\emptyset\}, \forall t \in [0,1]
\end{equation}
\noindent thus ensuring that the robot does not intersect with the displaced obstacles. 
\end{definition}

\section{Approach}
\label{sec:approach}
In this Section, we provide a systematic exposition of our approach, describing in detail the overlap and displacement stages. 
\subsection{Overlap Stage}
In this stage, the robot computes a trajectory minimizing a weighted sum of robot state cost, control cost and the overlap cost as defined in~(\ref{eq:overlap_cost}). 
The overlap cost $c^i(v^i)$ is a function of the measure of robot-obstacle overlap and is formulated based on the type of the constraint displacement problem. 
  
Let us consider a robot with a general dynamics model of the form
\begin{equation}
x_{k+1} = f(x_k,u_k) 
\label{eq:dynamic}
\end{equation}
\noindent where $x_{k+1}$, $x_k$ are the robot states at time $k+1$, $k$ respectively, and $u_k$ is the applied control at $k$. 
To formulate the overlap cost, we define a metric $\mathcal{L}(\bar{x},o^i)$ to measure the robot-obstacle overlap, where $\bar{x} = g(x_k)$ denotes the relevant robot states of interest. 
For example, for a mobile robot operating in a plane, $\bar{x}$ is the location of the robot whereas $x_k$ denotes its pose. 
If the robot does not overlap with the obstacle or just touches the obstacle, then $\mathcal{L}=0$. 
Maximum value of $\mathcal{L}$ is obtained when the robot completely overlaps with the obstacle. 
$\mathcal{L}$ is merely a function directly proportional to the robot-obstacle overlap and for a 3-D environment, we consider the 2-D projections or footprints of the robot and the obstacles. 
To compute $\mathcal{L}$, we thus assume the polygonal shaped obstacles and the robot to be bounded by circles of minimum area. 
This assumption provides an efficient approximation, and allows for a pertinent computation of $\mathcal{L}$ utilizing the distance between intersecting circles. 
For a robot of radius $r^r$ that overlaps an obstacle of radius $r^o$, $\mathcal{L}$ is defined as
\begin{equation}
   \mathcal{L}(\bar{x},o^i) =   (r^r+r^o) - \norm{x'-o}  
\label{collcond}
\end{equation} 
\noindent with $x'$, $o$ denoting the centers of the robot and obstacle circles, respectively, and $\norm{\cdot}$ denoting the $L_2$ norm, measuring the distance between the centers.
Therefore,~(\ref{collcond}) is thus the overlap cost $c^i(v^i)$ defined in~(\ref{eq:overlap_cost}). 
We note here that~(\ref{collcond}) is computed only when there is an overlap and,  otherwise, $\mathcal{L}=0$ as per the definition in~(\ref{eq:overlap_cost}). 
Furthermore, if two circles overlap or collide, then it holds that $\norm{x'-o} \leq (r^r+r^o)$. 
Thus $\mathcal{L}$ is essentially the collision condition, and it measures the extent of overlap when the robot and the obstacle intersect. 
We also note here that when a robot and an obstacle overlap, $\mathcal{L}$ is essentially the minimum distance by which the circle representing the obstacle is to be moved to achieve zero overlap -- the displacement is to be performed along the line joining robot's and obstacle's centers.
\begin{figure}[t]
  \subfloat[]{\includegraphics[trim=5 10 5 10,clip,scale=0.4]{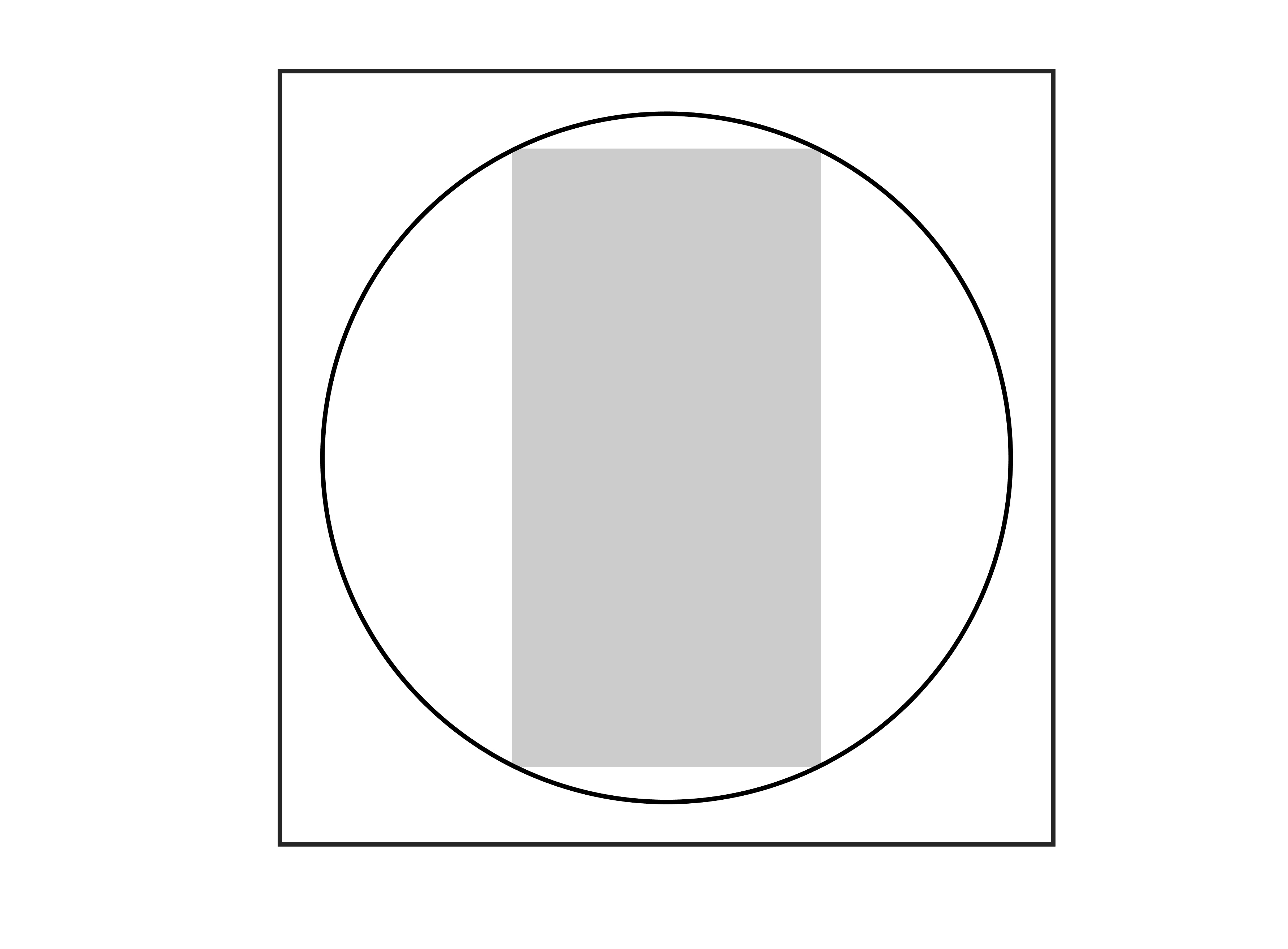}\label{fig:m1}}\hfill
  \subfloat[]{\includegraphics[trim=5 10 5 10,clip,scale=0.4]{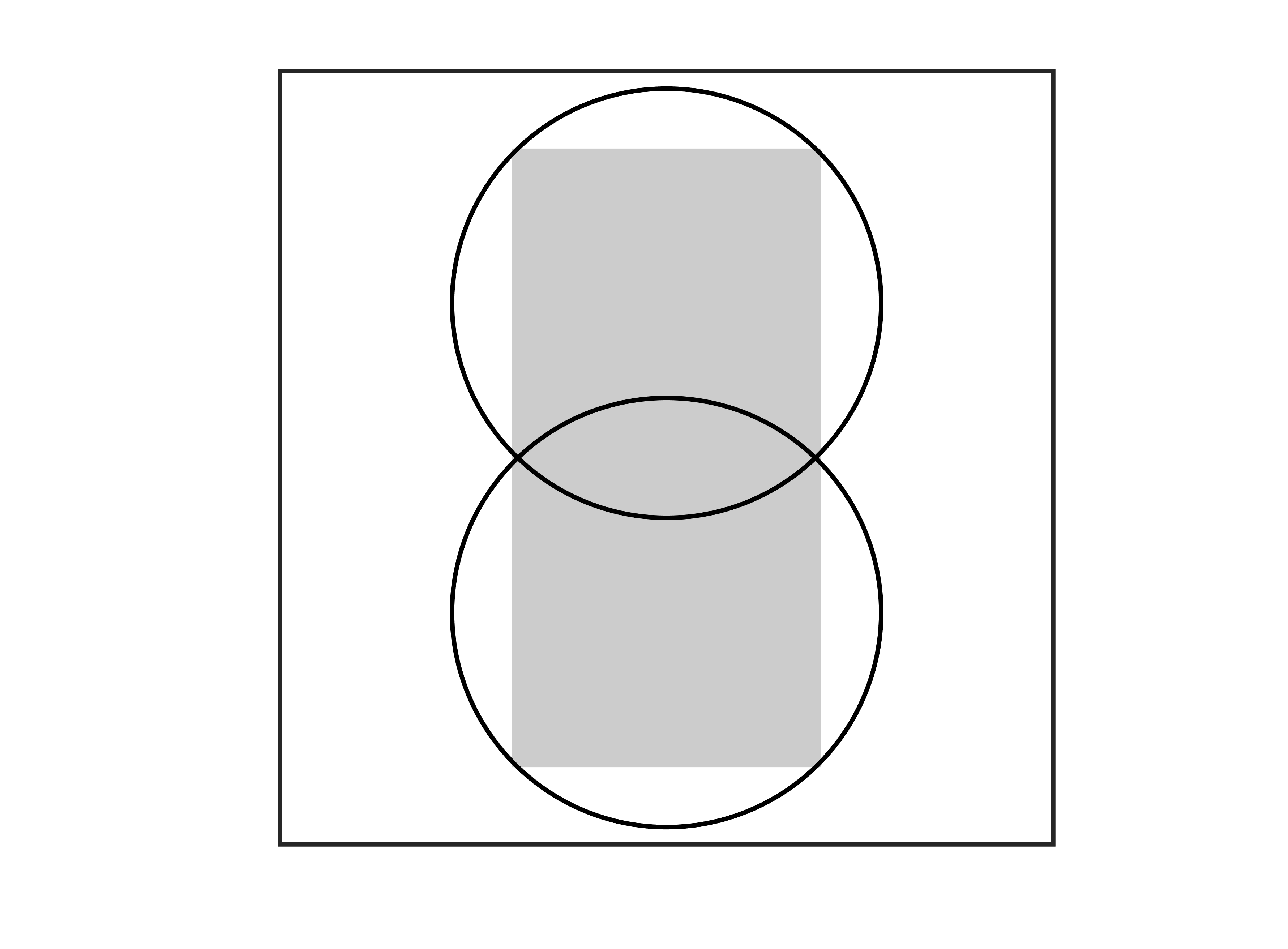}\label{fig:m2}}
\caption{(a) A rectangle (gray) represented using a single bounding circle. (b) Two circles used to bound the rectangle.}
\label{fig:multi}
\end{figure}

Generally, a robot and an obstacle (that is, their 2-D projections) may not be represented by single bounding circles, but by a number of minimum area enclosing circles.
Bounding a polygon using a single circle can give rise to overly conservative estimates, and to prevent this we use a combination of circles to compactly bound the polygon. 
Fig.~\ref{fig:multi} illustrates this point. 
$\mathcal{L}$ is thus computed for each pair of such robot-obstacle circles. 
This does not affect our problem definition since we are only concerned about the overall robot-obstacle overlaps, which is essentially captured by these robot-obstacle circular pairs. 

At each time instant $k$, the robot plans for $L$ look-ahead steps while minimizing an objective function
\begin{equation}
J_k = \sum\limits_{l=0}^{L-1} c_l(x_{k+l},\mathcal{O}) + c_L(x_{k+L})
\end{equation}
\noindent with $c_l$, $c_L$ being the cost for each look-ahead step and the terminal cost, respectively and defined as 
\begin{eqnarray}
c_l(x_{k+l},\mathcal{O}) = \norm{x_{k+l}}^2_{M_x} + \sum_{i=1}^n \norm{h(\mathcal{L}^i)}^2_{M_i} + \norm{\xi(u_{k+l})}^2_{M_u} \\
 c_L(x_{k+L}) = \norm{x_{k+L} - x^g}^2_{M_g} 
 \end{eqnarray}
\noindent where $\norm{x}_S = \sqrt{x^TSx}$ is the Mahalanobis norm, $\xi(\cdot)$ is a function that quantifies control usage and $M_x, M_i, M_u, M_g$ are weight matrices. 
The optimization problem of the overlap stage can now be formally stated as 
\begin{mini!}|s|[1]
{ }{ \sum\limits_{l=0}^{L-1}\left[ \norm{x_{k+l}}^2_{M_x} + \sum_{i=1}^n \norm{h(\mathcal{L}^i)}^2_{M_i} + \norm{\xi(u_{k+l})}^2_{M_u} \right]  + \norm{x_{k+L} - x^g}^2_{M_g} }{\label{eq:cost_fn}}{}
  \addConstraint{x_0}{=x^s}
  \addConstraint{x_{k+l}}{=f(x_{k+l-1},u_{k+l-1})}
    \addConstraint{u_{k+l}}{\in U}{\label{eq:controls}}
      \end{mini!}
\noindent where~(\ref{eq:controls}) constraints the control actions to lie within the feasible set of control inputs. 
The term $\norm{h(\mathcal{L}^i)}^2_{M_i}$ penalizes the robot-obstacle overlaps and the optimization returns the control actions $u_0,...,u_{T-1}$. 
Executing the synthesized control results in a trajectory $x_0,\ldots,x_{T}$ that overlaps with the obstacles to be displaced. 
The function $h(\cdot)$ varies according to the constraint displacement problem. 
For example, for an MCD problem, $h(\cdot)$ is an identity function, that is, $h(\mathcal{L}^i) = \mathcal{L}^i$, since we are minimizing the overall obstacle displacements. 
The constraint $x_T = x^g$ is omitted since this is a soft constraint that depends on the robot model and control limits. The overlap stage is summarized below:\\
\textbf{Overlap Stage:}\\
1. Input: $x^s$, $x^g$, $\mathcal{O}$, $M_x, M_i, M_u, M_g, \mathcal{D}$.\\
2. Perform the optimization in~\eqref{eq:cost_fn}.\\
3. Output: $u_0,...,u_{T-1}$.\\\\
Now that we have computed the controls and hence the robot trajectory, the overlapping obstacles need to be moved so that the robot can safely execute the computed trajectory.   

\subsection{Displacement Stage}
In this stage we execute the controls computed in the previous stage, that is, $u_0,...,u_{T-1}$ to simulate $x_0, \ldots, x_T$, which  gives a trajectory with robot-obstacle overlaps. 
While doing so, we consider the actual polygonal shaped robot and obstacles. 
It is now sufficient to displace the overlapping obstacles.

The problem can be stated as follows: 
\textit{Given two overlapping or intersecting polygons, how can we displace one of the polygon (the obstacle) to a new location where it has zero overlap with the other polygon (the robot)?} 
Irrespective of the type of the problem (MCD or MCR) we are interested in minimizing the displacement of obstacles and therefore the new location is obtained while reducing a certain displacement metric. 
We note here that in general an obstacle may not be in overlap with a single polygonal robot, but different robot polygons, since the robot trajectory is continuous. 
As a result, the displacement should ensure that the respective obstacle do not overlap with each of the robot polygons along its trajectory. 
Below, we will delineate the displacement process considering a pair of polygons. 
Extension to multiple polygon overlaps readily follows.

\noindent \textbf{Circle-circle overlap.} 
If the robot and the obstacles are circular in shape (or spheres in 3-D), we have already seen before that $\mathcal{L}$ is the minimum distance the obstacle circle is to be moved along the line joining the robot and obstacle centers so as to obtain zero overlap.

\noindent \textbf{Circle-polygon overlap.} 
We first begin with a circle and a line segment overlap. 
Let us consider a line segment and a circle in the plane. 
The line segment either lies outside the circle, overlaps with the circle at a single point or overlaps with the circle at two distinct points. 
We suppose the line segment to be long enough so that the case of line segment being entirely inside the circle is avoided. 
We will now derive the condition for a line segment to not overlap with a circle.

\begin{figure}[t]
\subfloat[]{\includegraphics[trim=10 40 5 10,clip,scale=0.3]{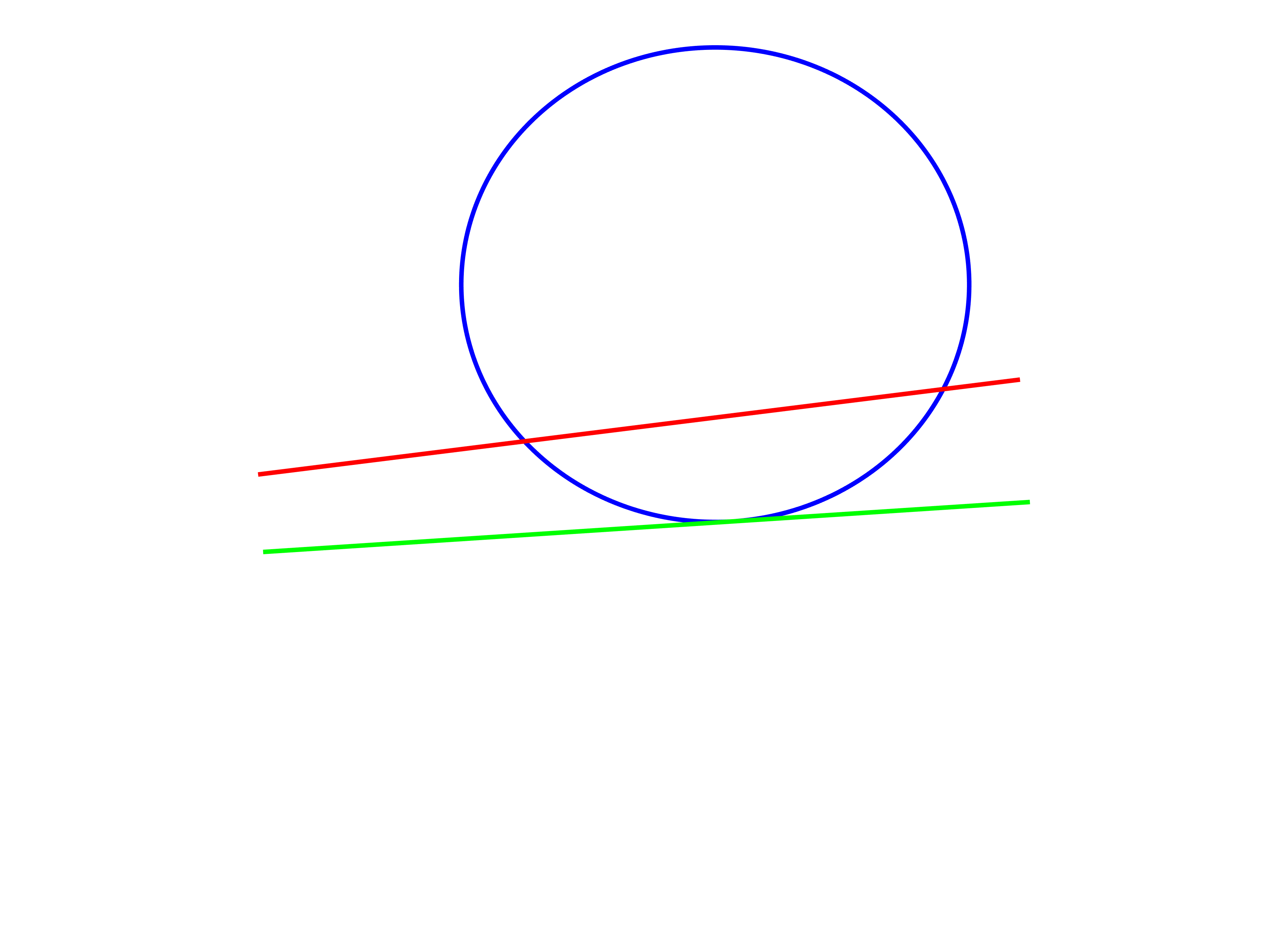}\label{fig:l1}}
\subfloat[]{\includegraphics[trim=5 10 5 10,clip,scale=0.3]{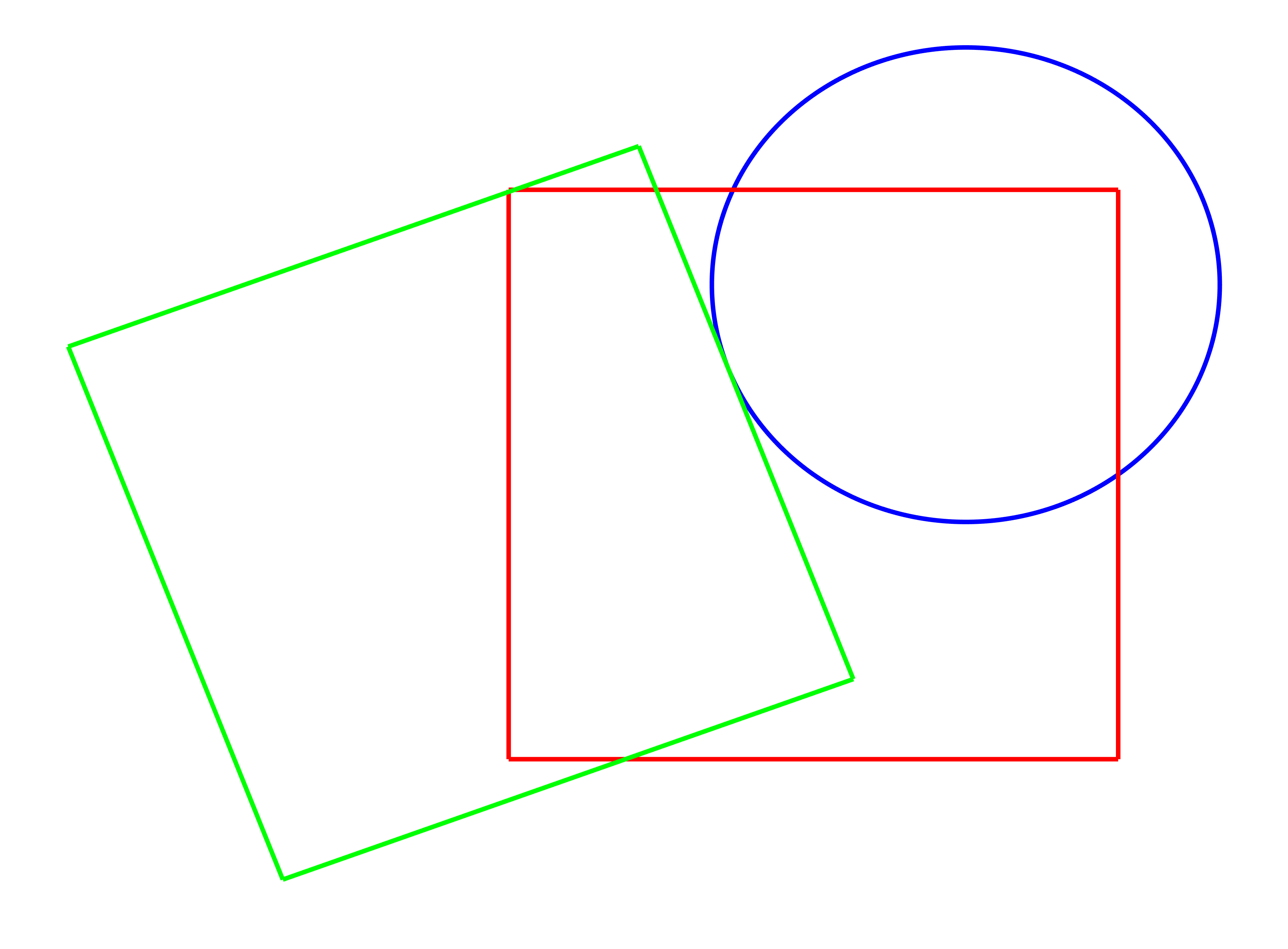}\label{fig:l2}}
\subfloat[]{\includegraphics[trim=5 10 5 2,clip,scale=0.3]{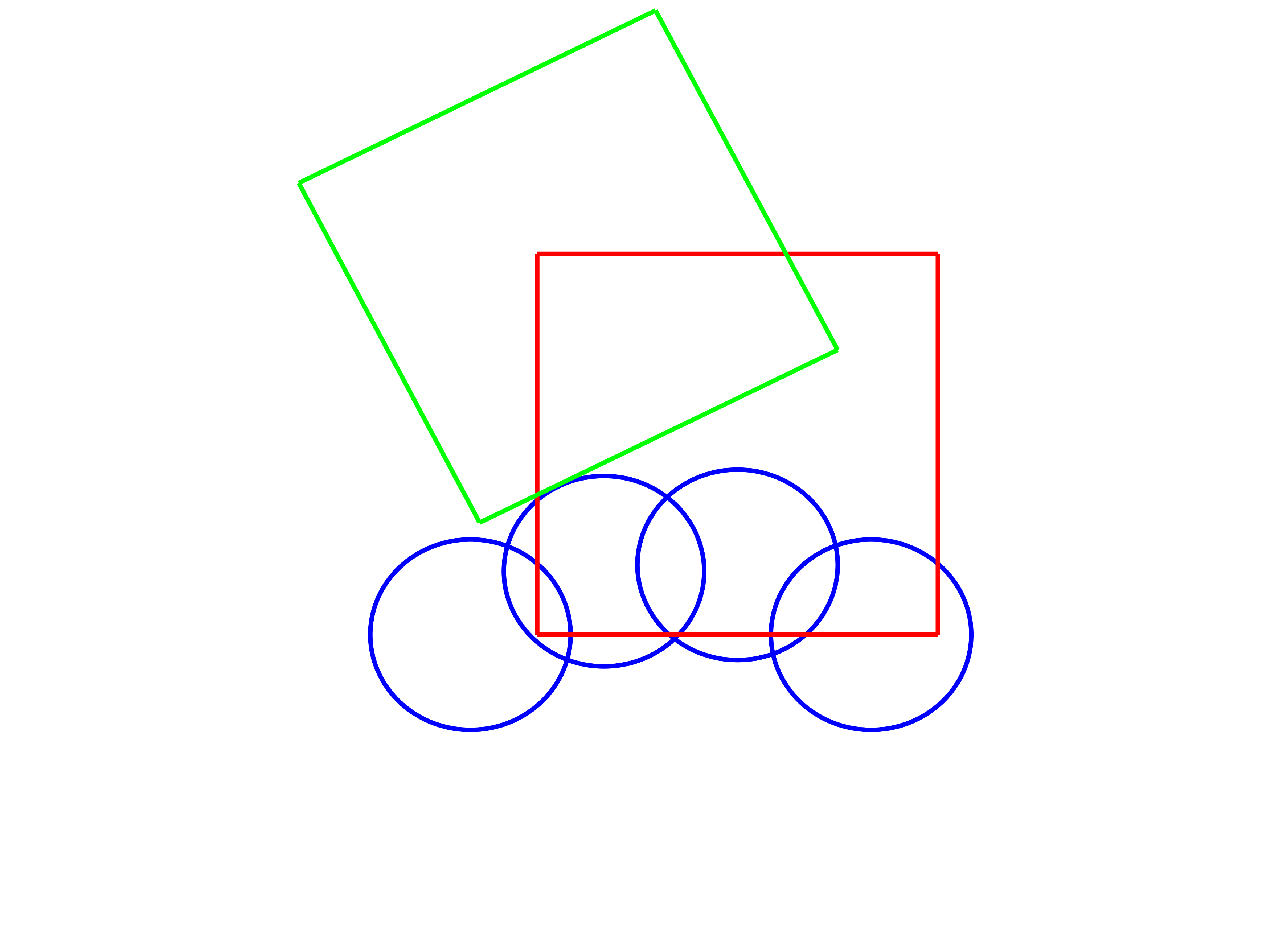}\label{fig:l3}}
\caption{
Illustration of circle-polygon overlap. 
(a) The overlapped line in red is displaced to a new location. 
(b) The green square is the displaced location with no overlap with the circle. 
(c) Multiple circles intersect with a square. 
The optimization in~(\ref{eq:cpdisp}) computes a zero overlap location (in green) for the square.
}
\label{fig:circle-poly}
\end{figure}

Let us consider a circle with center $(x,y)$ and radius $r$, and a line segment with endpoints $(x1,y1)$ and $(x2,y2)$. 
Any point on the line segment can be represented by $\left(tx1 + (1-t)x2, ty1 + (1-t)y2\right)$, where $0\leq t \leq 1$. 
Let us assume that the line segment overlaps with the circle. 
The point or points of overlap can be found by substituting $\left(tx1 + (1-t)x2, ty1 + (1-t)y2\right)$ in the equation of the circle and solving for the parameter $t$. 
We thus have
\begin{equation}
(tx1 + (1-t)x2 - x)^2 + (ty1 + (1-t)y2 - y)^2 = r^2
\end{equation}
\noindent Upon expanding and rearranging the above equation, we get the following quadratic equation in terms of $t$
\begin{equation}
l^2t^2 + 2\left((x1-x2)(x2-x) + (y1-y2)(y2-y)\right)t + x2^2 + x^2 + y2^2 + y^2 - r^2 = 0
\end{equation} 
\noindent where $l = \sqrt{(x1-x2)^2 + (y1-y2)^2}$ is the length of the line segment. 
If the line segment does not overlap, then the above equation has no solution and therefore the discriminant $D$ of the quadratic equation should be negative. 
Evaluating $D$ and rearranging, we get the following condition for no overlap
\begin{equation}
-\left((x1-x2)(y2-y) + (y1-y2)(x2-x)\right) + l^2r^2 < 0
\label{eq:circle-poly}
\end{equation}
Let us now come back to our problem. 
If a line segment with endpoints $(x1,y1)$ and $(x2,y2)$ overlaps with the circle, we would like to find a new location for the line segment so that there is no overlap. 
We thus require to find new endpoints $(d1,d2)$ and $(d3,d4)$ for the line segments satisfying~(\ref{eq:circle-poly}) such that the line segment no longer overlaps with the circle. 
Furthermore, to preserve the length of the line segment, $l =\sqrt{(d1-d3)^2 + (d2-d4)^2}=\sqrt{(x1-x2)^2 + (y1-y2)^2}$ should also be satisfied. 
The required displacement is thus formulated as the following nonlinear optimization problem
\begin{mini!}|s|[1]
{ }{ (x1-d1)^2 + (y1-d2)^2 + (x2-d3)^2 + (y2-d4)^2}{\label{eq:cpdisp}}{}
\addConstraint{-\left((x1-x2)(y2-y) + (y1-y2)(x2-x)\right) + l^2r^2}{<0}
\addConstraint{(d1-d3)^2 + (d2-d4)^2}{=l^2}
\end{mini!}
\noindent where~(\ref{eq:cpdisp}a) is the objective or the minimized displacement. 
The objective function is not restrictive to the one defined here, but it may take several other forms that capture the displacement. For example, we may also minimize the change in the midpoint of the initial and the displaced line segments. 
Note that the above method for computing new end points for the line segment also encodes rotational displacements and~(\ref{eq:cpdisp}) tends to capture the overall rotation and transnational displacements. 
Furthermore, the formulation readily extends to a polygon since it is formed by joining different line segments together. 
Additional constraints need to be added to ensure that the polygonal shape is maintained. 
For example, in the case of a rectangle, the additional constraints include maintaining the length of the two diagonals. Fig.~\ref{fig:circle-poly} illustrates this exposition. 
In Fig.~\ref{fig:l1}, a line segment (in red) overlaps a circle (in blue). 
The new displaced location without overlap is computed employing the optimization in~(\ref{eq:cpdisp}) and is shown in green.
Fig.~\ref{fig:l2} shows a similar scenario for a circle and a square. 
In Fig.~\ref{fig:l3}, multiple circles overlap with the square and the square is moved to a new location to obtain zero overlap. 
The different circles can be envisioned as robot configurations along a trajectory. 

\begin{figure}[t]
\subfloat[]{\includegraphics[trim=10 40 5 10,clip,scale=0.3]{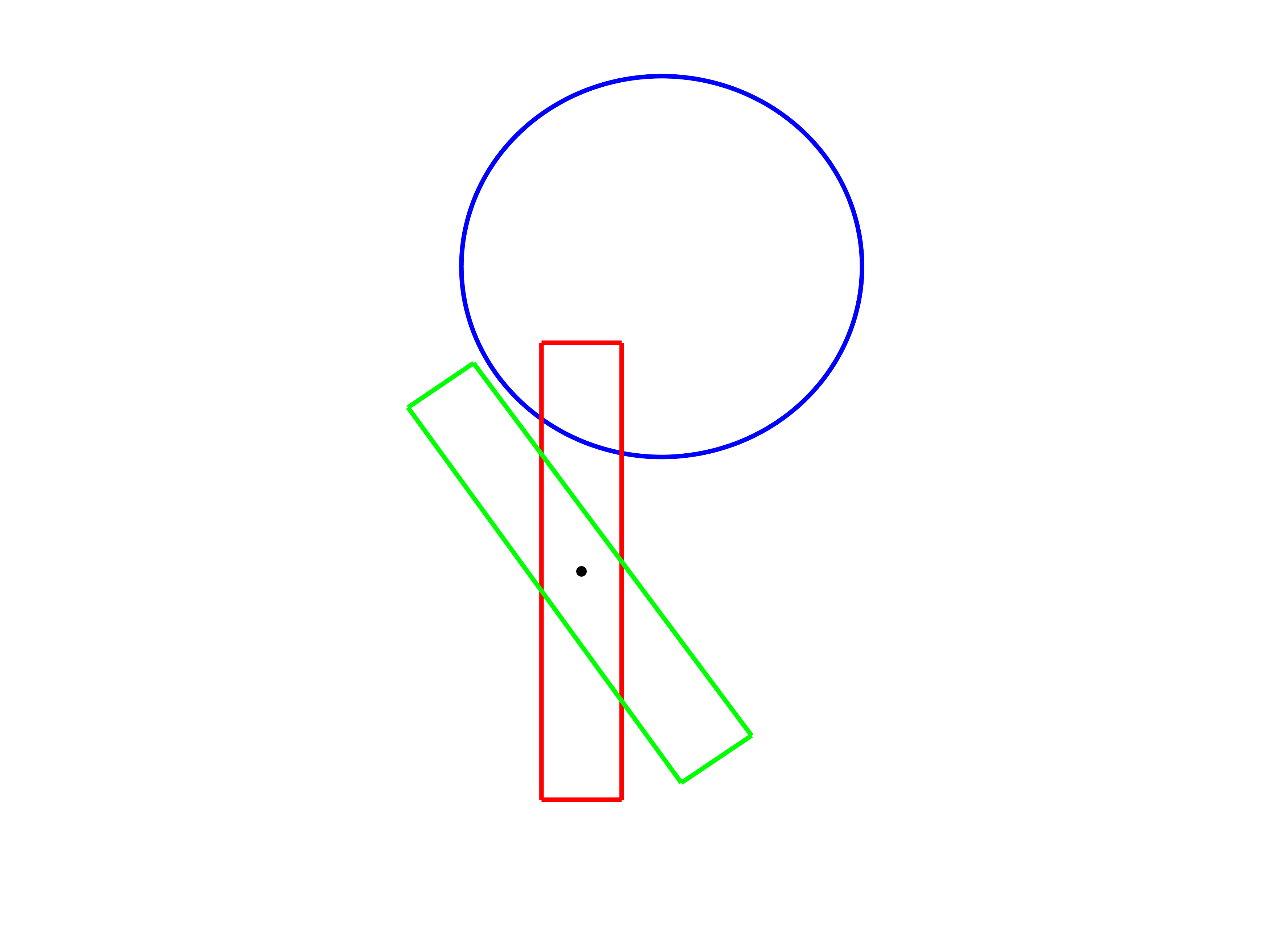}\label{fig:il1}}
\subfloat[]{\includegraphics[trim=5 10 5 10,clip,scale=0.3]{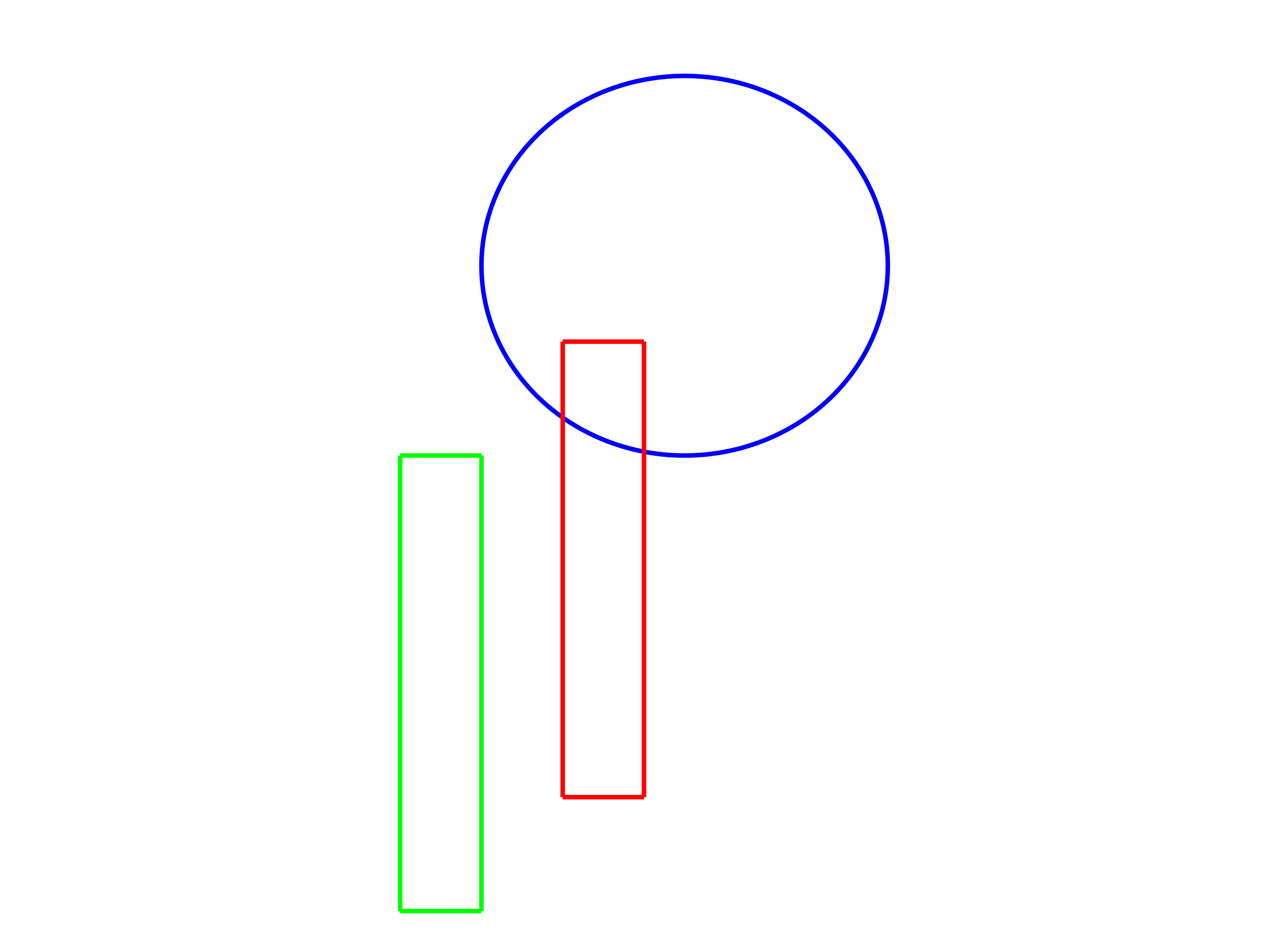}\label{fig:il2}}
\caption{
Example of pure rotations and translations. 
Additional constraints are added in~(\ref{eq:cpdisp}) to obtain the desired results. 
(a) A rectangular rod (in red) than can perform only pure rotations. 
The displaced rectangle is shown in green. 
(b) A rectangle than can only perform transnational motions. 
The overlapping rectangle is translated to a new location (green) with no overlap.
}
\label{fig:il}
\end{figure}

As noted before, the formulation in~(\ref{eq:cpdisp}) encodes both rotations and translations. 
Under scenarios in which only either rotations or translations are possible, the formulation can be slightly modified to obtain the desired results. 
For example, in the case of pure rotations of a rectangle, apart from the constraints in~(\ref{eq:cpdisp}), additional constraints are to be added to ensure that the mean of diagonally opposite coordinates is equal to the coordinates of midpoint point of the rectangle. 
Similarly, for pure translations of a rectangle, the difference between consecutive $x-$coordinates and the difference between consecutive $y-$coordinates should remain constant. 
An illustration is shown in Fig.~\ref{fig:il}. 

\noindent \textbf{Polygon-polygon overlap.} 
We first consider the condition for two line segments to not intersect. 
Let us consider two line segments with end points $(x1,y1)$, $(x2,y2)$ and $(x3,y3)$, $(x4,y4)$, respectively. 
Any point on these line segments can be represented by $\left(tx1 + (1-t)x2, ty1 + (1-t)y2\right)$ and $(sx3 + (1-s)x4, sy3 +$ $ (1-s)y4)$, respectively, with $0\leq t \leq 1$ and $0\leq s \leq 1$. 
If the two lines intersect, we can equate the intersection point (or points, if the line segments are the same) on these line segments and we obtain a linear equation in terms of $t$, $s$ as
\begin{equation}
\begin{bmatrix}
(x1-x2) & \ -(x3-x4) \\ 
(y1-y2) & \ -(y3-y4) \\
\end{bmatrix}
\begin{bmatrix}
t \\ 
s  \\
\end{bmatrix}=
\begin{bmatrix}
x4-x2  \\ 
y4-y2  \\
\end{bmatrix}
\label{eq:poly-poly}
\end{equation} 
The above linear equation is of the form $Ax = b$ and has no solution if the inverse of $A$ does not exist.  
The condition of non existence of the inverse gives either no solution with the two line segments being parallel to each other or infinite solution with the line segments being on top of each other, that is, complete overlap. 
We ignore these solutions where the two line segments are parallel to each other ($det(A)=0$), and instead we look at the solutions with $t \notin [0,1]$ and $s \notin [0,1]$. 
Solving for $t$, $s$ from~(\ref{eq:poly-poly}), we get
\begin{equation}
\begin{split}
t &= \frac{-(y3-y4)(x4-x2) + (x3-x4)(y4-y2)}{-(x1-x2)(y3-y4) + (y1-y2)(x3-x4)}\\ 
s &= \frac{-(y1-y2)(x4-x2) + (x1-x2)(y4-y2)}{-(x1-x2)(y3-y4) + (y1-y2)(x3-x4)} \\
\end{split}
\label{eq:ts}
\end{equation}
For no intersection, we need $t \notin [0,1]$ and $s \notin [0,1]$. 
We use the following condition to ensure no intersection 
\begin{equation}
\begin{split}
1/t & <= 1 \\ 
1/s & <= 1 \\
\end{split}
\label{eq:pol-pol}
\end{equation}

\begin{figure}[t!]
\centering
\includegraphics[scale=0.45]{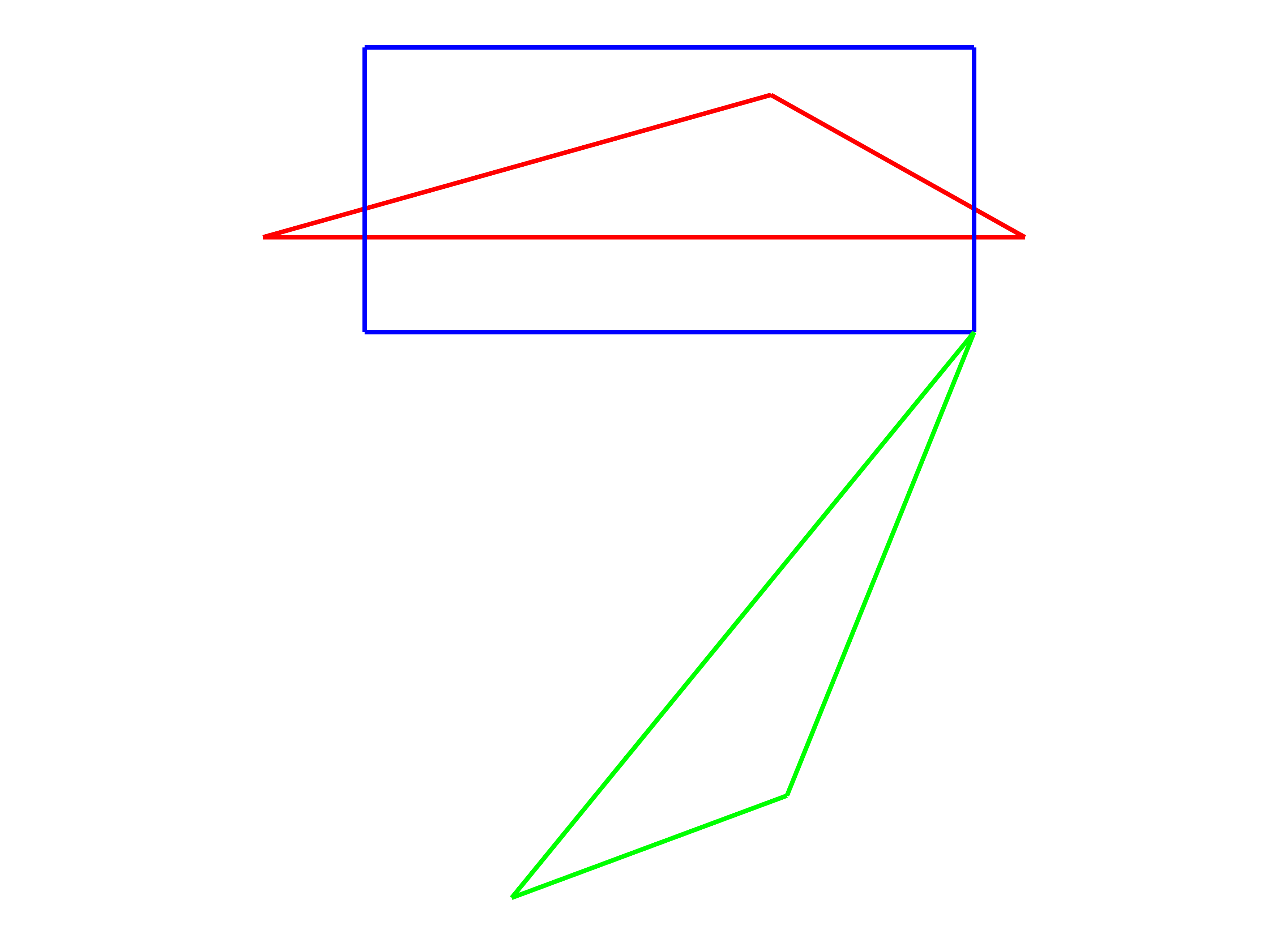}
\caption{
A rectangle (in blue) and a triangle (in red) are intersecting initially. 
A new location (in green) is found for the movable triangle with no intersection with the rectangle.
}
\label{fig:rect-tri}
\end{figure} 

Let us assume that the line segment with endpoints $(x1,y1)$, $(x2,y2)$ is the one to be displaced. 
Let the endpoints of this line segment after displacement be $(d1,d2)$, $(d3,d4)$. 
This displaced line segment must therefore satisfy~(\ref{eq:pol-pol}) and $l = \sqrt{(x1-x2)^2 + (y1-y2)^2} = \sqrt{(d1-d3)^2 + (d2-d4)^2}$. 
The optimization can now be stated as
\begin{mini!}|s|[1]
{ }{(x1-d1)^2 + (y1-d2)^2 + (x2-d3)^2 + (y2-d4)^2}{\label{eq:ppdisp}}{}
\addConstraint{1/t}{<=1}{\label{eq:ppdisp1}}
\addConstraint{1/s}{<=1}{\label{eq:ppdisp2}}
\addConstraint{(d1-d3)^2 + (d2-d4)^2}{=l^2}
\end{mini!}
\noindent
We add an $\epsilon \ll 1$ to the numerators of $t$ and $s$ in~(\ref{eq:ts}) so as to avoid division by zero during the numerical computation of~(\ref{eq:ppdisp1}) and~(\ref{eq:ppdisp2}). 
As it has been argued before, the formulation readily extends to polygon intersections. 
Fig.~\ref{fig:rect-tri} shows a rectangle and triangle overlapping. 
Optimization in~(\ref{eq:ppdisp}) is performed to displace the triangle to a new location (in green in the figure) with zero overlap with the rectangle. 
\begin{figure}[]
\subfloat[]{\includegraphics[trim=52 110 38 101,clip,scale=0.5]{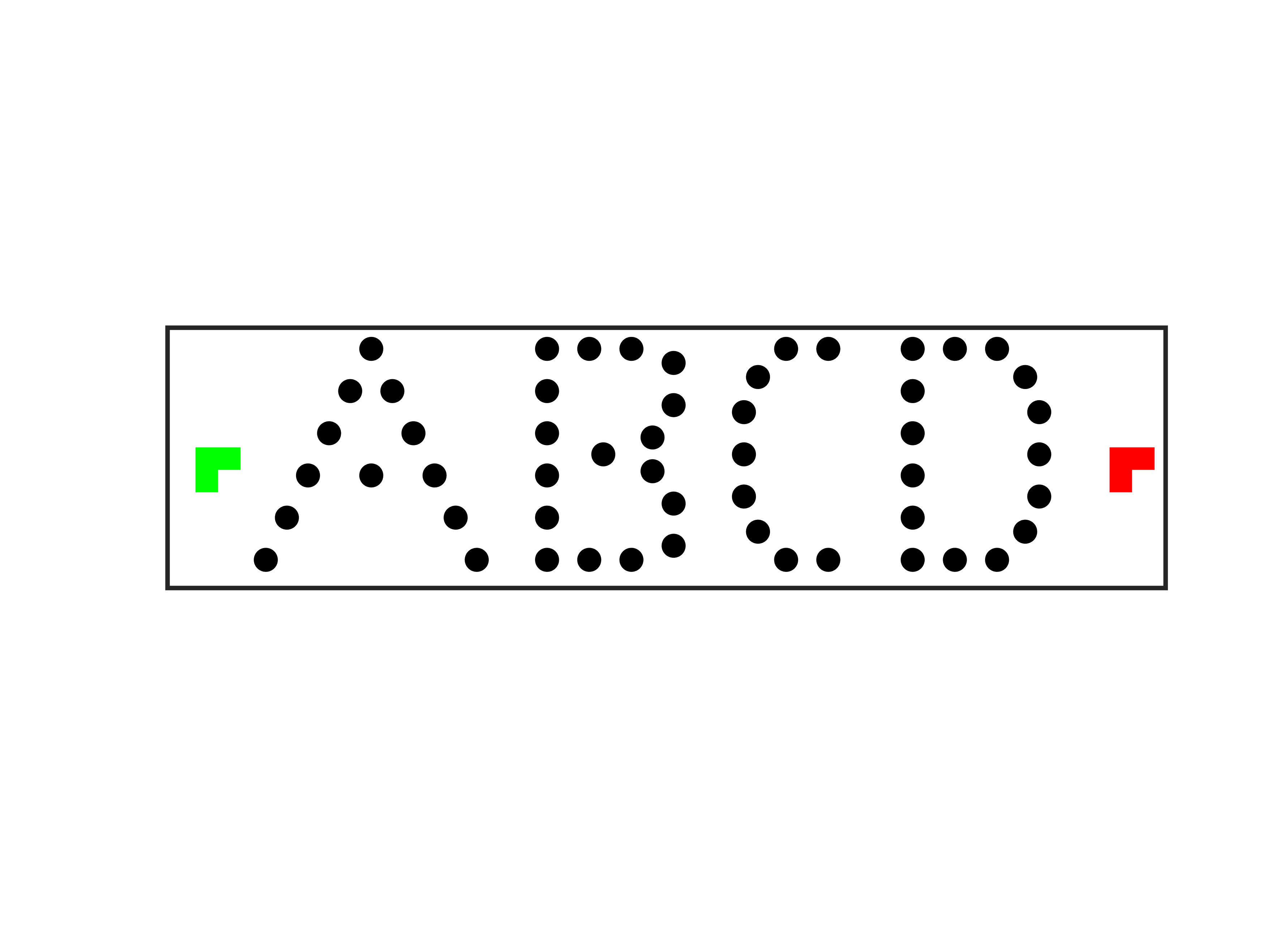}\label{fig:ex0}}\hfill
\subfloat[]{\includegraphics[trim=52 110 38 101,clip,scale=0.5]{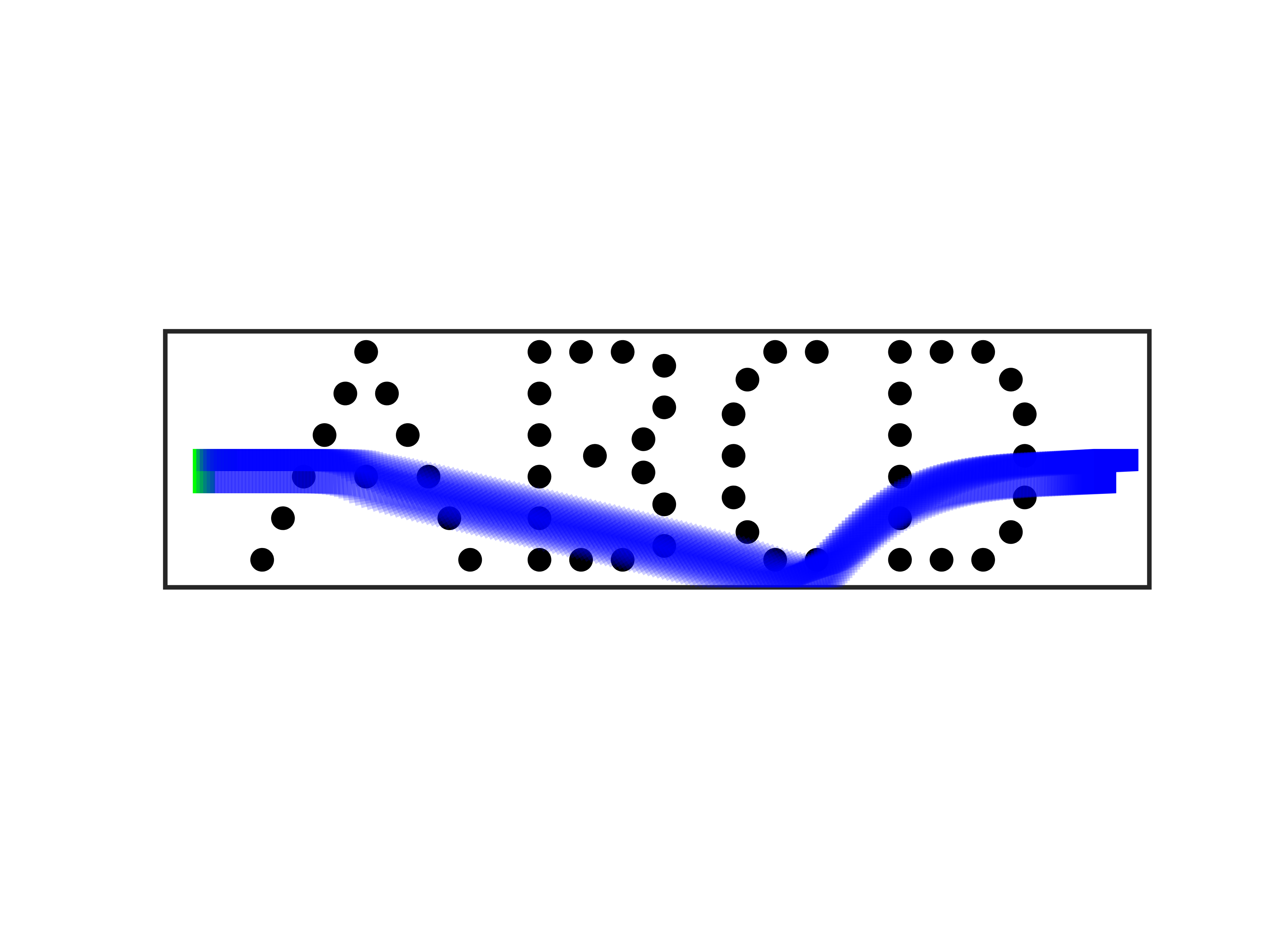}\label{fig:ex1}}\hfill
\subfloat[]{\includegraphics[trim=52 110 38 101,clip,scale=0.5]{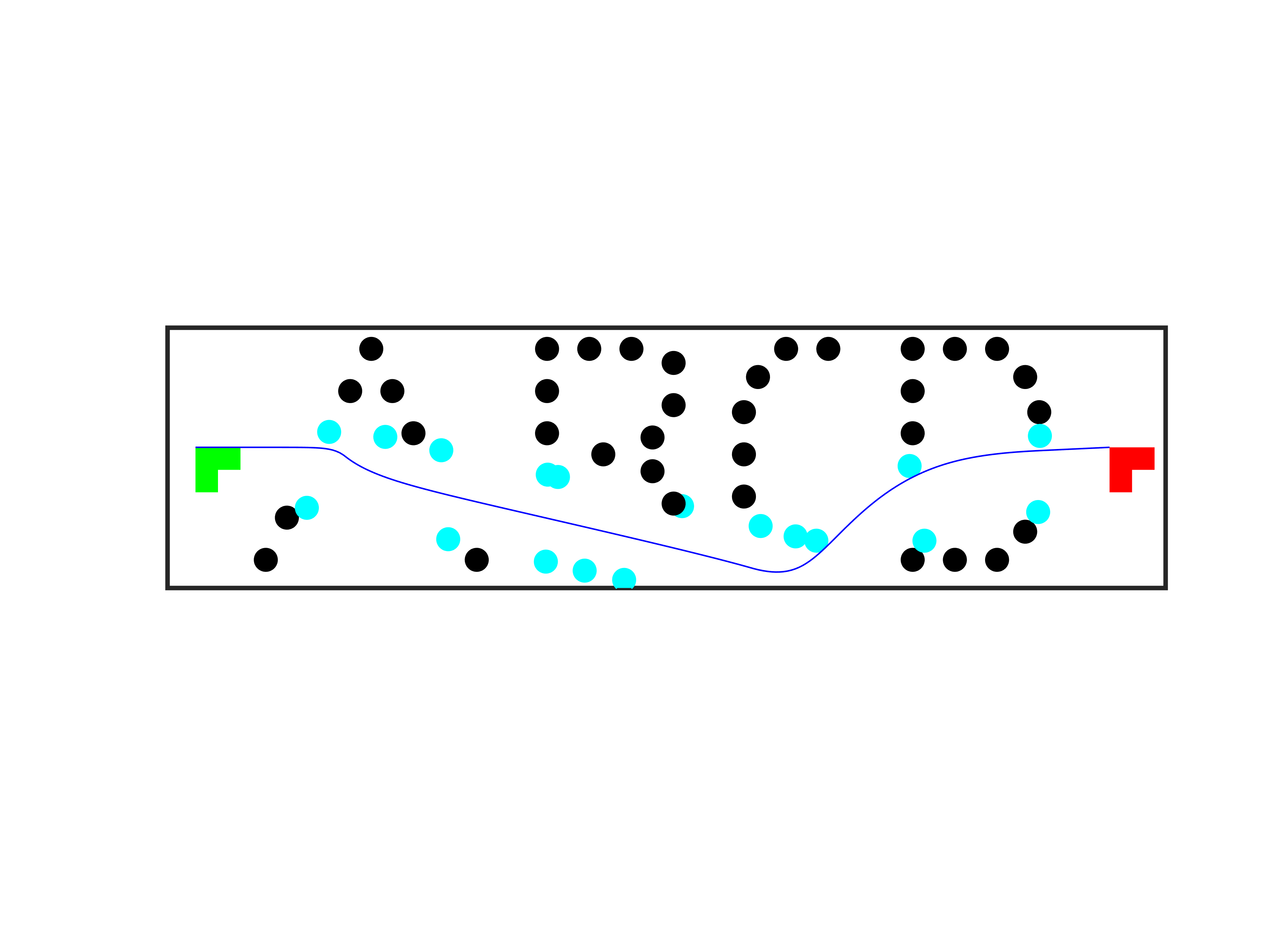}\label{fig:ex2}}\hfill
\caption{
Illustration of the constraint displacement problem. 
(a) An L-shaped robot must move from left to right in the presence of movable circular obstacles. 
The robot can translate and rotate in the plane. 
(b) The simulated robot trajectory using the control actions synthesized in the overlap stage. 
(c) The overlapping obstacles are displaced resulting in a path with no collision.
}
\label{fig:ex}
\end{figure}

An illustrative example for this stage is shown in Fig.~\ref{fig:ex}. 
In this setup, an L-shaped robot in the \textit{ABCD} domain needs to reach the right hand side of the map by moving the circular obstacles (see Fig.\ref{fig:ex0}). 
The overlap stage results in the control commands $u_0,\ldots,u_T$, which upon simulation lead to a trajectory with obstacle overlaps as shown in Fig.~\ref{fig:ex1}. 
Once the required obstacle displacements are computed, the obstacles are cleared to give a collision free trajectory as can be seen in Fig.~\ref{fig:ex2}.

We note here that the overlap removal problem finds applications in several areas outside the application scope discussed herein. 
These include geometric representation of graphs and networks~\cite{hayashi1998ISGD}, or archaeological applications~\cite{van2017CGF}. However, these approaches often find non overlapping rectangles while maintaining an orthogonal order, that is, the order of the rectangles with respect to the $x$ and the $y$ axis do not change and tend to be computationally intensive. 
Our approach is general with no restriction on the nature or order of displacements.

\section{Implementation and Implications}
\label{sec:implementation}

In the previous Section, we have presented the constraint displacement problem by dividing it into two sub-problems -- the overlap stage and the displacement stage. 
The overlap stage essentially optimizes a function of robot overlap with the obstacles. 
For the minimum constraint displacement problem, in which the robot finds a feasible path while minimizing displacement magnitudes, the overall obstacle overlap is to be minimized. 
This is because an obstacle needs to be displaced only if there is an overlap with the planned robot trajectory.
Therefore, minimizing the overall overlap with all the movable obstacles during planning leads to minimum displacement (magnitude) during execution. 
The corresponding objective function is as given in~(\ref{eq:cost_fn}) with $h(\cdot)$ being an identity function. 
For an MCR problem, the unique obstacle overlaps should be minimized as we look for solutions with the minimum number of displaced obstacles (or minimum obstacles to be removed). 
A similar reasoning could be used to formulate the objective function for manipulation under clutter. 
An objective function for the MCR problem can be found in Section~\ref{sec:results}. 
In the case of NAMO additional aspects such as the work done could be incorporated. 
This may involve prioritizing overlaps at the polygonal edges or contact points, for example to facilitate a better grasping of the object by the robot.

\begin{figure}[t]
\subfloat[]{\includegraphics[trim=10 40 5 10,clip,scale=0.4]{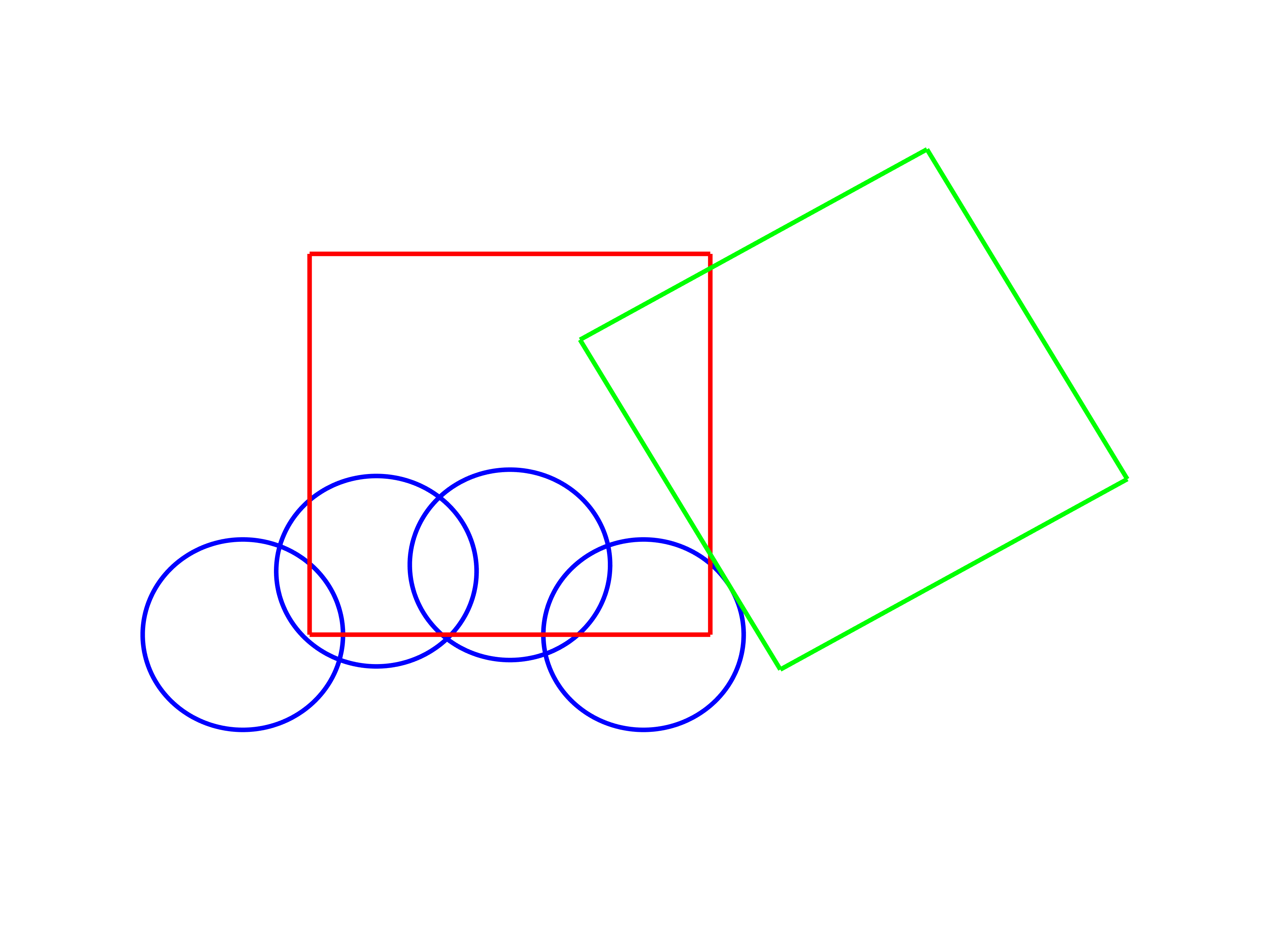}\label{fig:impl1}}\hfill
\subfloat[]{\includegraphics[trim=0 10 5 10,clip,scale=0.4]{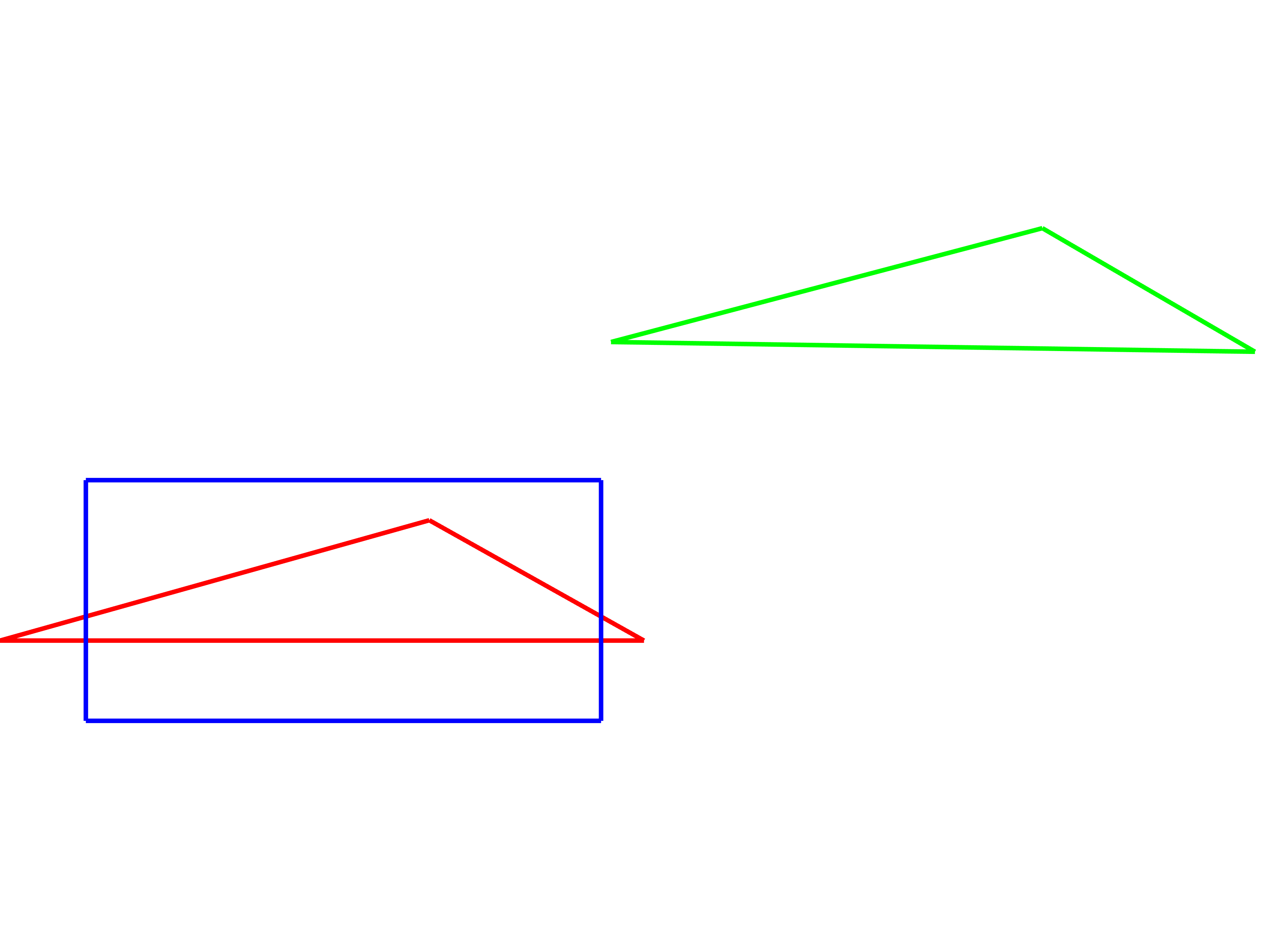}\label{fig:impl2}}
\caption{
Alternative solutions to the scenarios in Fig.~\ref{fig:l3} and Fig.~\ref{fig:rect-tri}, respectively, when different starting points are used.
}
\label{fig:impl}
\end{figure}

To implement the overlap stage, we employ the FORCESPRO~\cite{FORCESPro,FORCESNlp} solver, which generates optimized nonlinear model predictive control (MPC) code. 
In order to perform the displacement stage optimization, we employ the nonlinear optimization function \textsc{fmincon}~\cite{MATLABR2021a} with the interior point algorithm~\cite{byrd1999JOO}. 
The interior point algorithm is used since the models in~(\ref{eq:cpdisp}) and~(\ref{eq:ppdisp}) are non-convex. 
This approach proceeds iteratively by starting from the interior of the feasible set and hence solicit an initial feasible point. 
Initial points for the displacement stage are easily obtained by displacing the polygon in the $x$ or $y$ direction so that there is no overlap. 
For example, in the case of a circle-rectangle overlap, an initial set of feasible points for the rectangle vertices are obtained by displacing the $x$ or $y$ coordinates (direction depends on the nature of the overlap) of the rectangle by an amount $\Delta$. 
A safe value for $\Delta$ is the diameter of the circle. 
Furthermore, if we solicit different solutions, if they exist, they can be obtained by using $\Delta = \Delta \pm \delta$, where $\delta > 0$. 
However, a global optimum cannot be theoretically guaranteed for the \textsc{fmincon} optimization and the approach tend to return local minima. 
Though fast locally optimal solutions are obtained, the solutions are greatly affected by the initial and starting points.
Fig.\ref{fig:impl1} and Fig.~\ref{fig:impl2} illustrate different solutions to the optimization problem in Fig.~\ref{fig:l3} and Fig.~\ref{fig:rect-tri}, respectively, when different initial points are provided as input.

It should be noted that the work considered in this paper is an offline motion planning approach providing us with some leverage on  planning time. 
It could be argued that the optimization~\eqref{eq:cost_fn} of the overlap stage may be carried out at once for the entire time step. 
We note here that both MCD and MCR do not exhibit the optimal substructure property~\cite{cormen2009book} of dynamic programming, meaning that optimal solutions to sub-problems are not necessarily optimal (a consequence of the problems being NP-hard). 
Consequently, optimization over the entire time step can be computationally expensive. 
Therefore, the MPC approach provides a very reasonable approximation.

Additionally, for the circle-polygon overlap, we have ignored the solution set $t \notin [0,1]$ and $s \notin [0,1]$ from the optimization constraints in~(\ref{eq:cpdisp}), and for the polygon-polygon overlap, we have omitted the constraints corresponding to $det(A)=0$ (line segments being parallel). 
These assumptions allow us to simplify the optimization problems in~(\ref{eq:cpdisp}) and~(\ref{eq:ppdisp}). 
We note here that when a human moves around through a cluttered space, he or she rearranges or push the objects around to make way. 
Most often, the human consciously or subconsciously performs an \textit{optimization}, for example, in terms of the number of objects to push or finding the shortest path. However, as the human navigates quickly through the cluttered space, most often he or she is not concerned about globally optimal solutions, but locally optimal solutions that can be computed quickly through his thought process as soon as the map of the environment is registered. 
This justifies our simplification of the constraints and finding locally optimal solutions.

\section{Evaluation}
\label{sec:results}
\begin{figure}[]
  \subfloat[]{\includegraphics[trim=52 110 38 101,clip,scale=0.5]{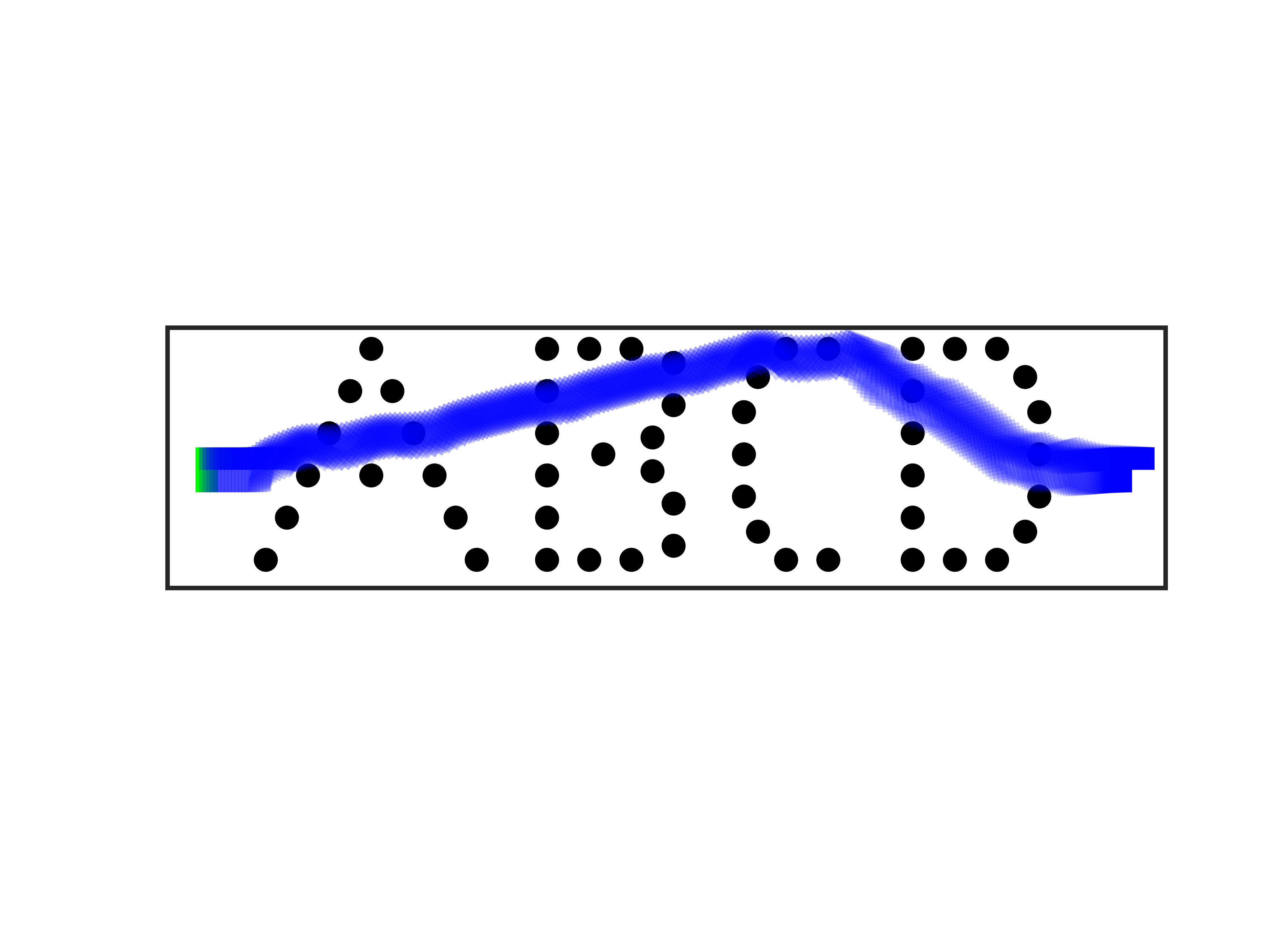}\label{fig:L1}}\hfill
  \subfloat[]{\includegraphics[trim=52 110 38 101,clip,scale=0.5]{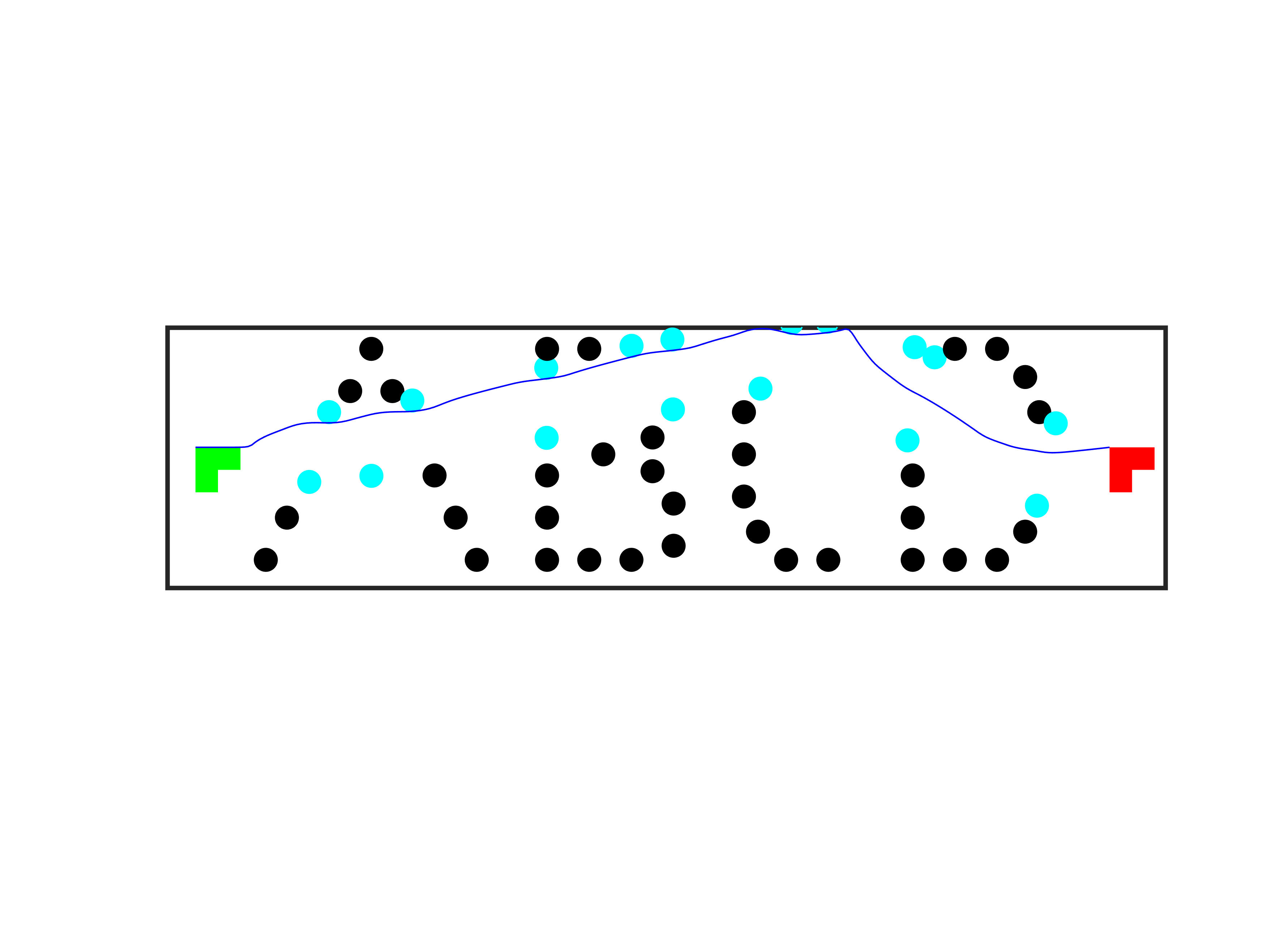}\label{fig:L2}}\hfill
  \subfloat[]{\includegraphics[trim=52 110 38 101,clip,scale=0.5]{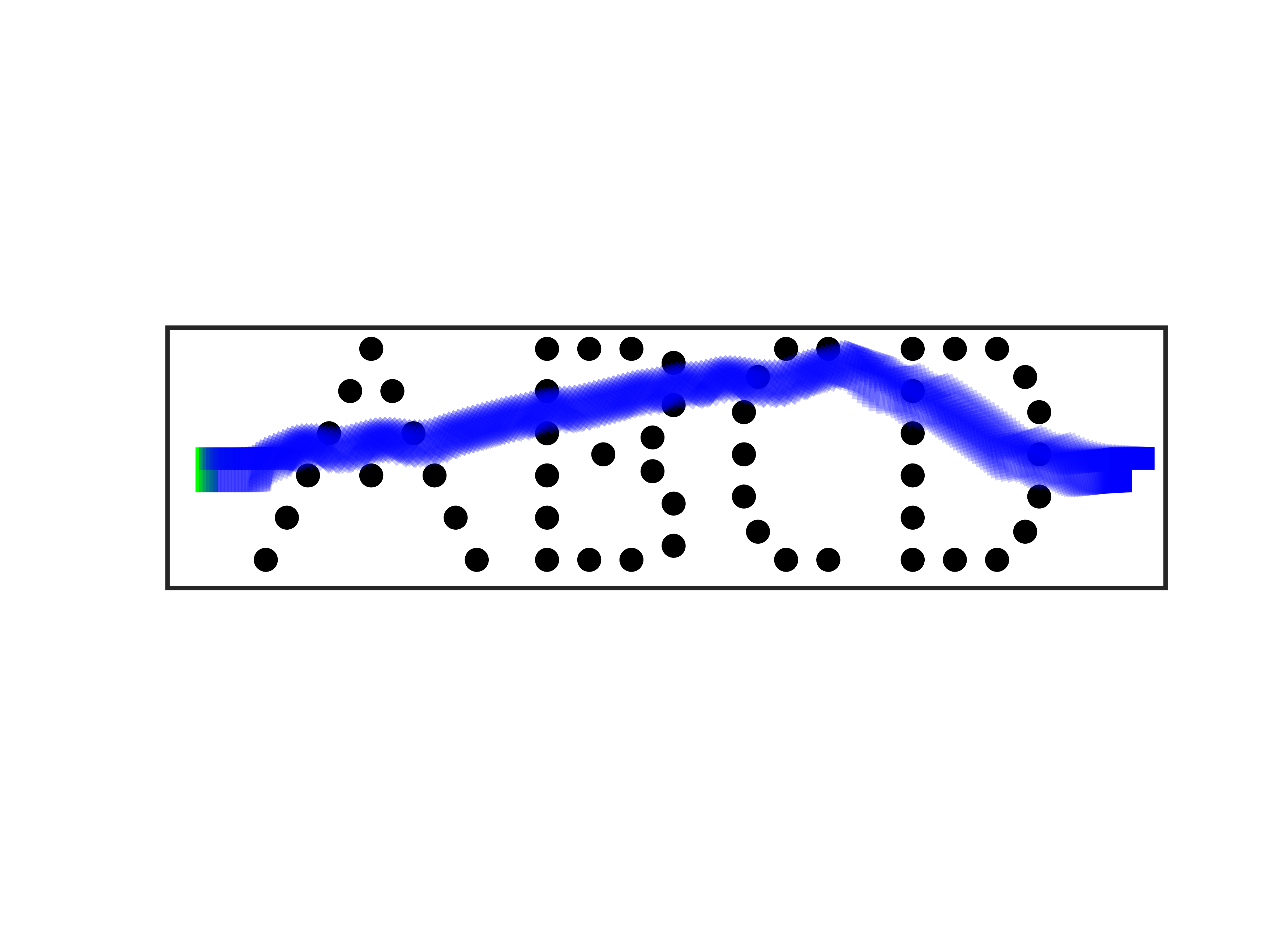}\label{fig:L3}}\hfill
  \subfloat[]{\includegraphics[trim=52 110 38 101,clip,scale=0.5]{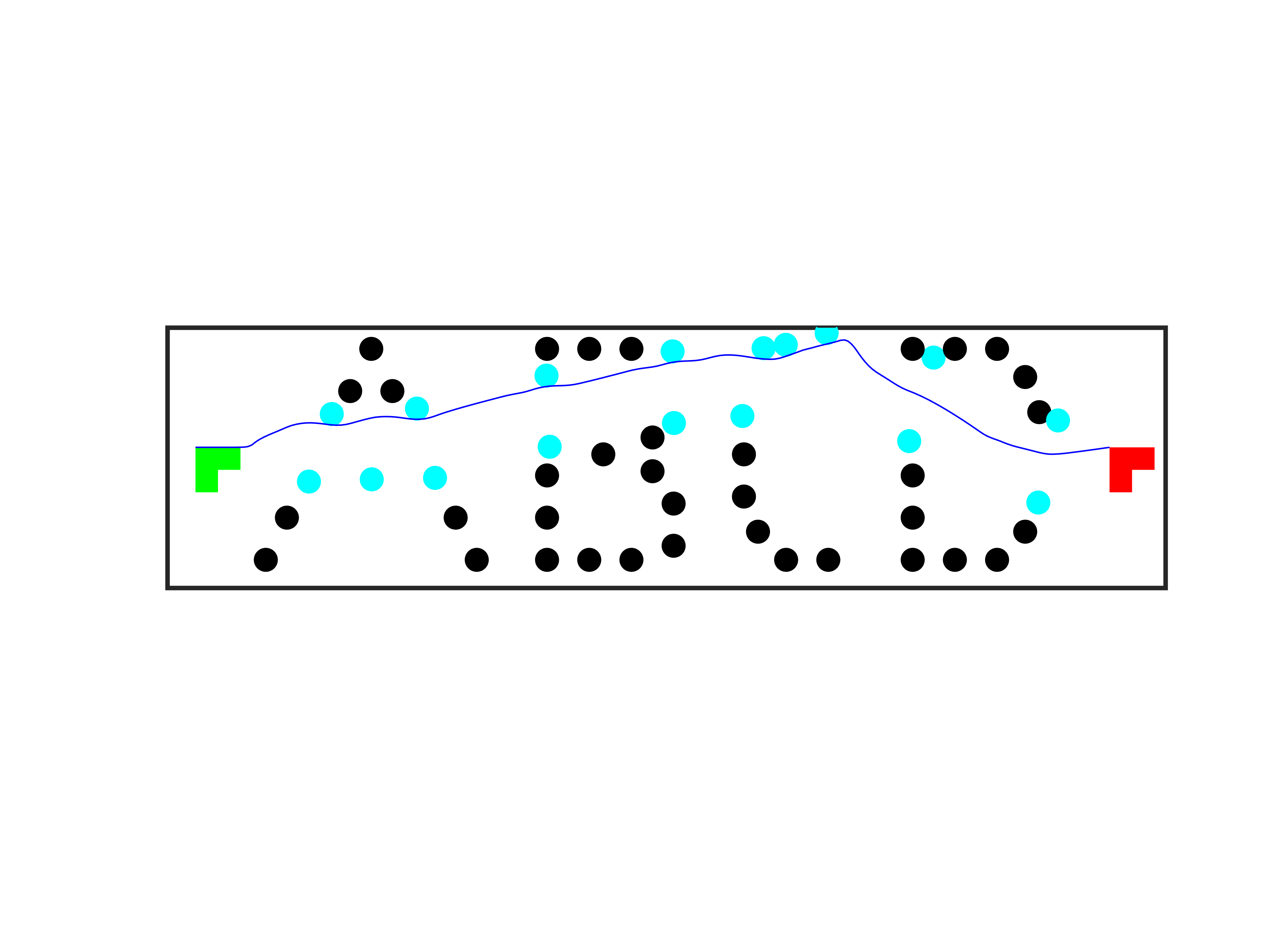}\label{fig:L4}}\hfill
  \subfloat[]{\includegraphics[trim=52 110 38 101,clip,scale=0.5]{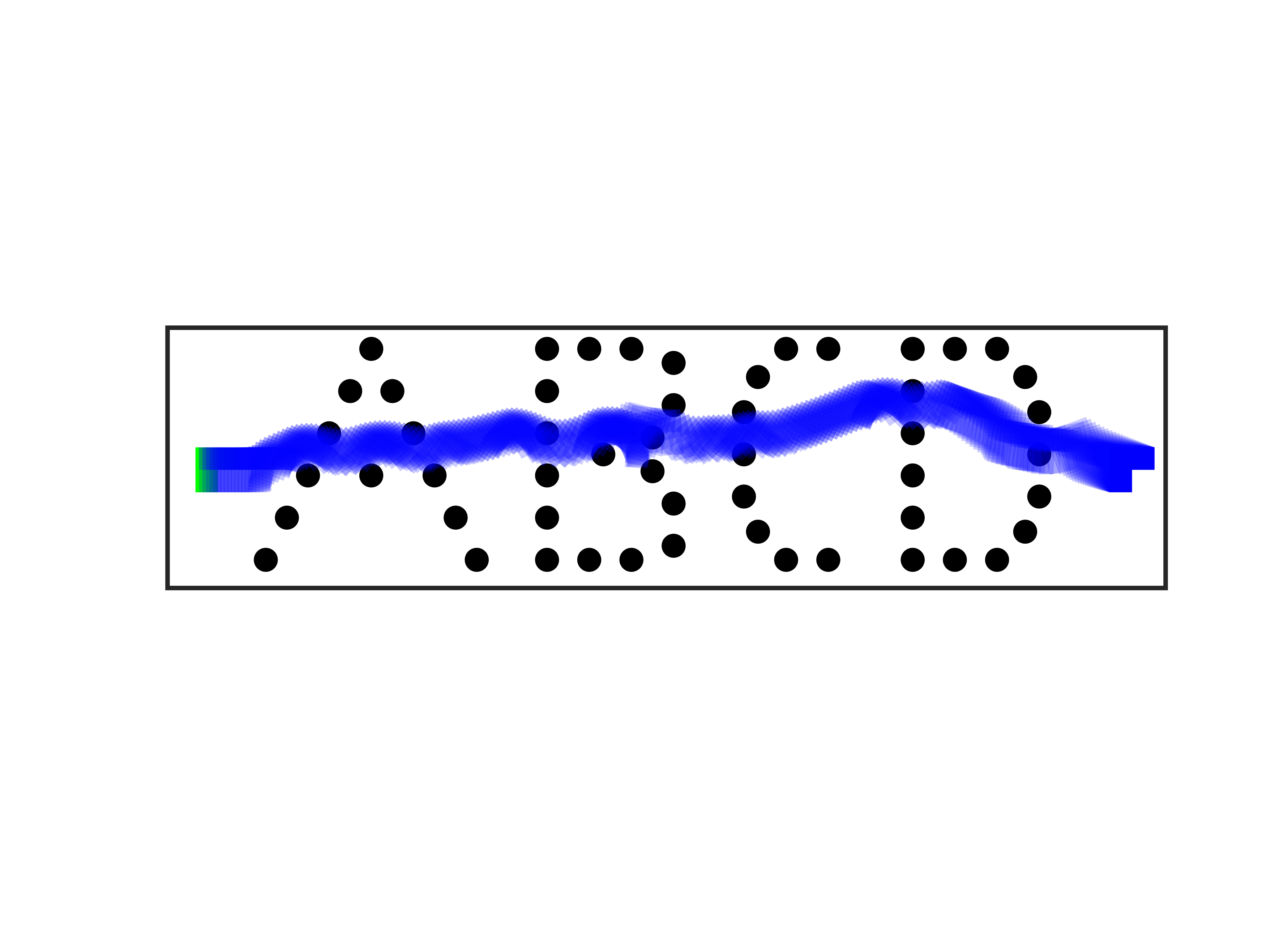}\label{fig:L5}}\hfill
  \subfloat[]{\includegraphics[trim=52 110 38 101,clip,scale=0.5]{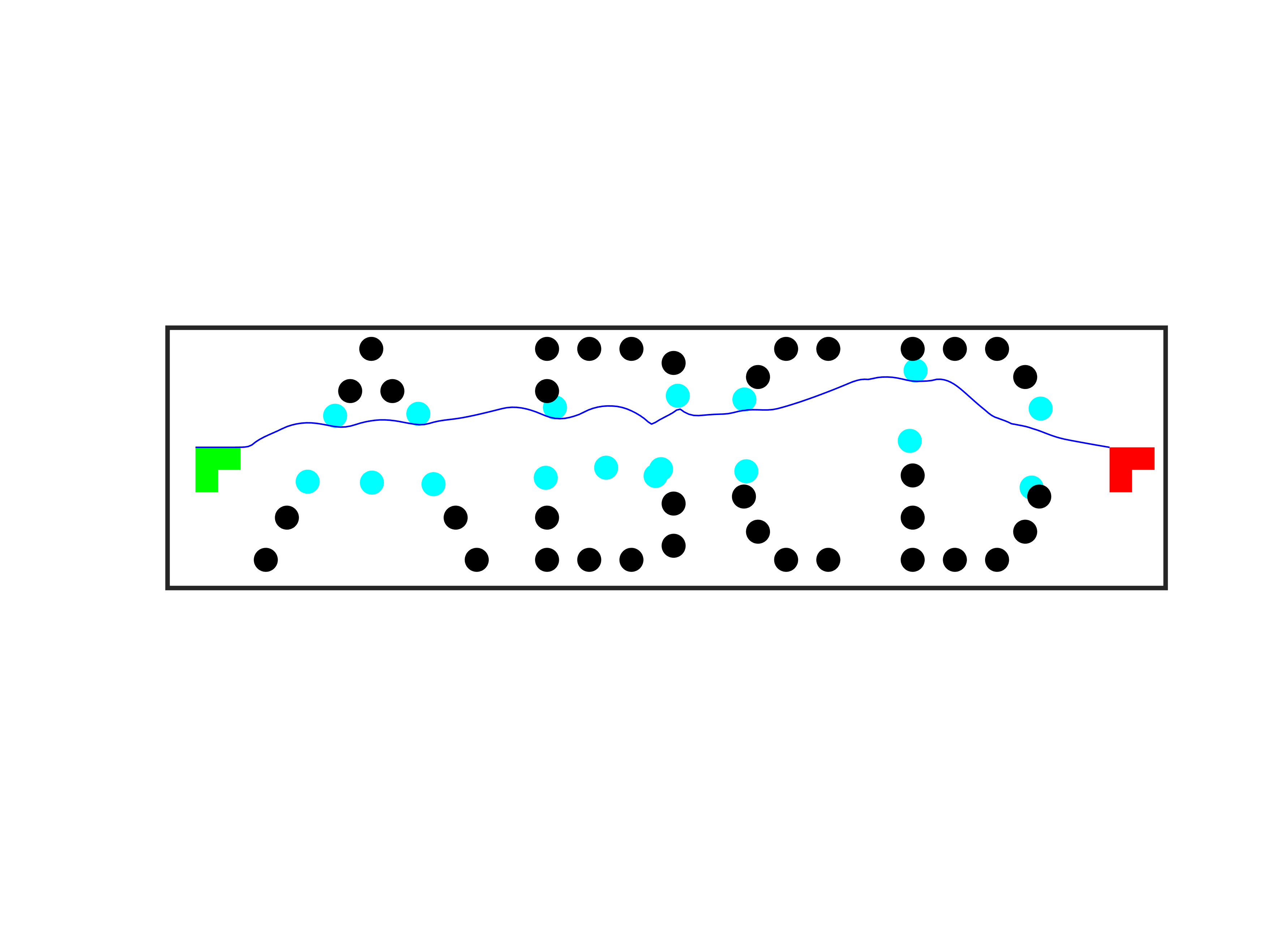}\label{fig:L6}}\hfill
   \subfloat[]{\includegraphics[trim=52 110 38 101,clip,scale=0.5]{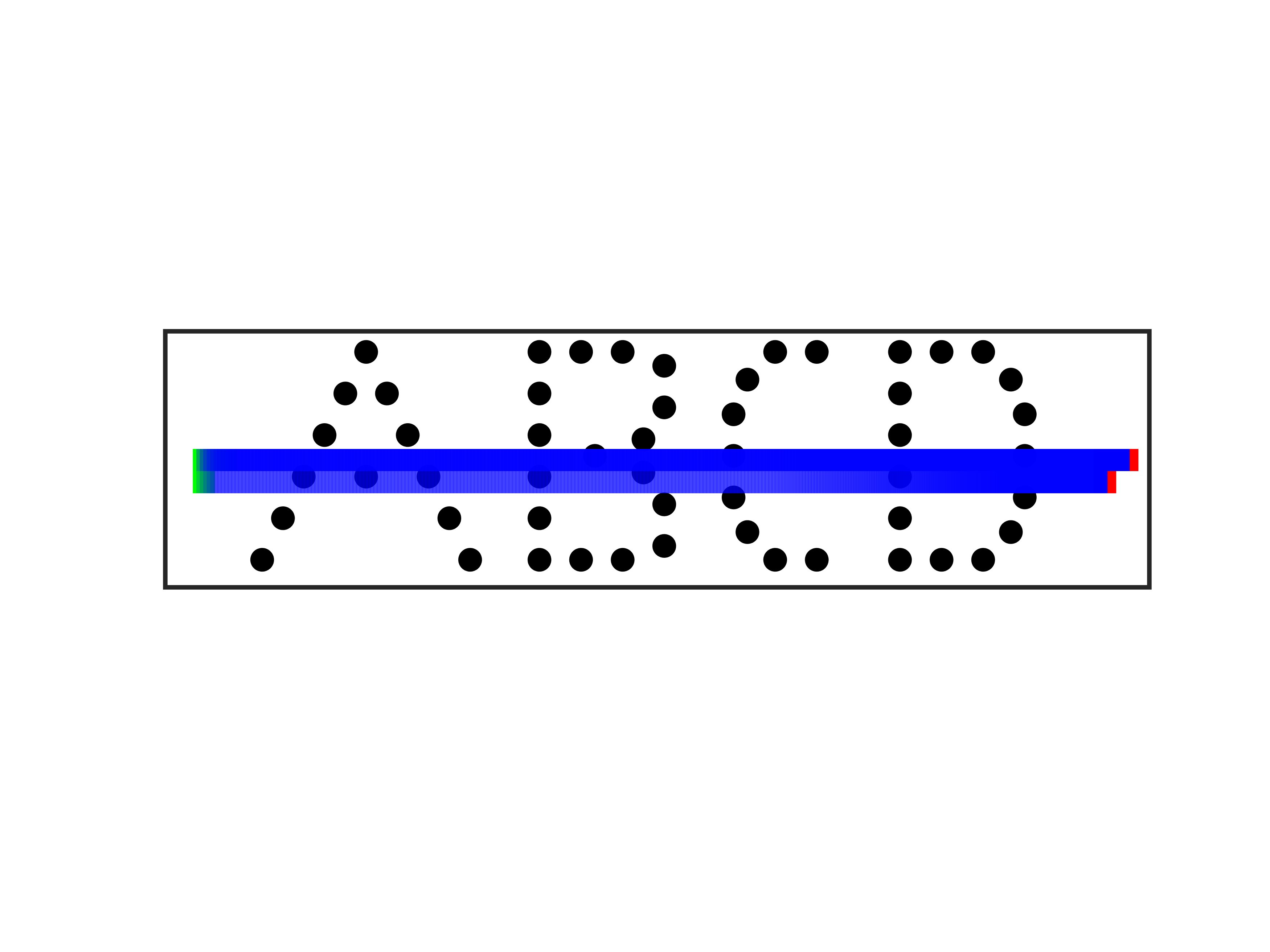}\label{fig:L7}}\hfill
  \subfloat[]{\includegraphics[trim=52 110 38 101,clip,scale=0.5]{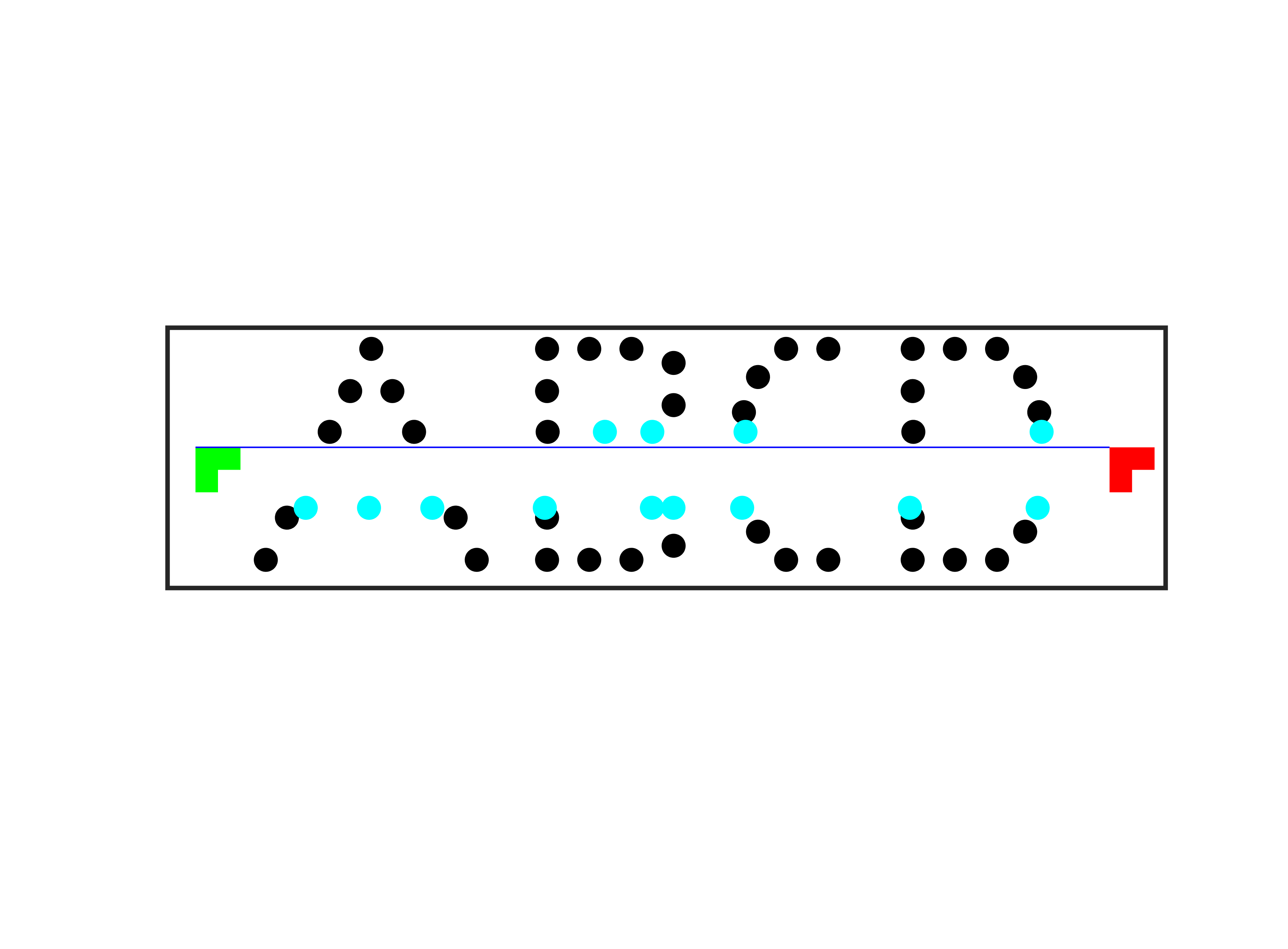}\label{fig:L8}}\hfill
  \subfloat[]{\includegraphics[trim=52 110 38 101,clip,scale=0.5]{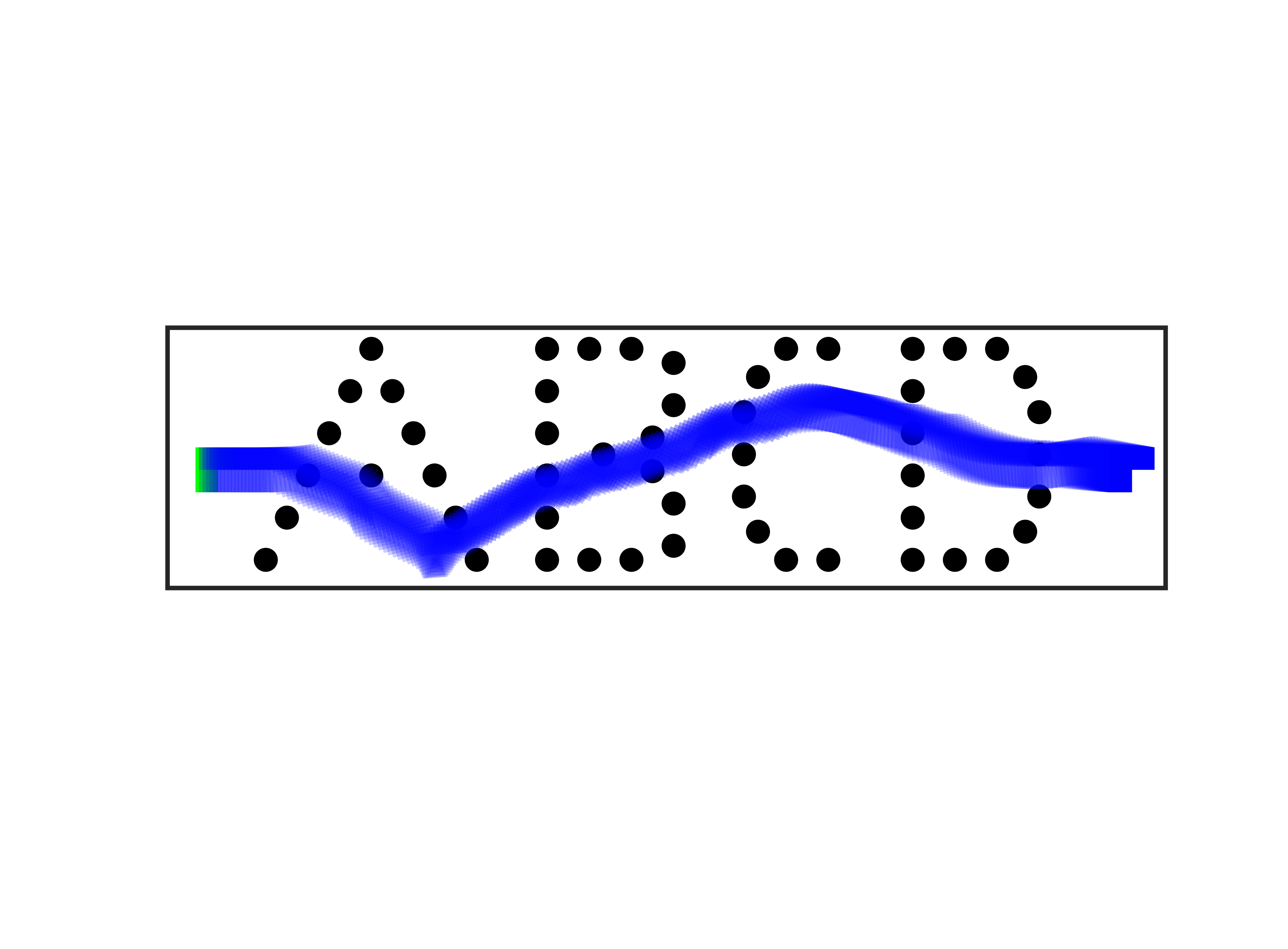}\label{fig:L9}}\hfill
  \subfloat[]{\includegraphics[trim=52 110 38 101,clip,scale=0.5]{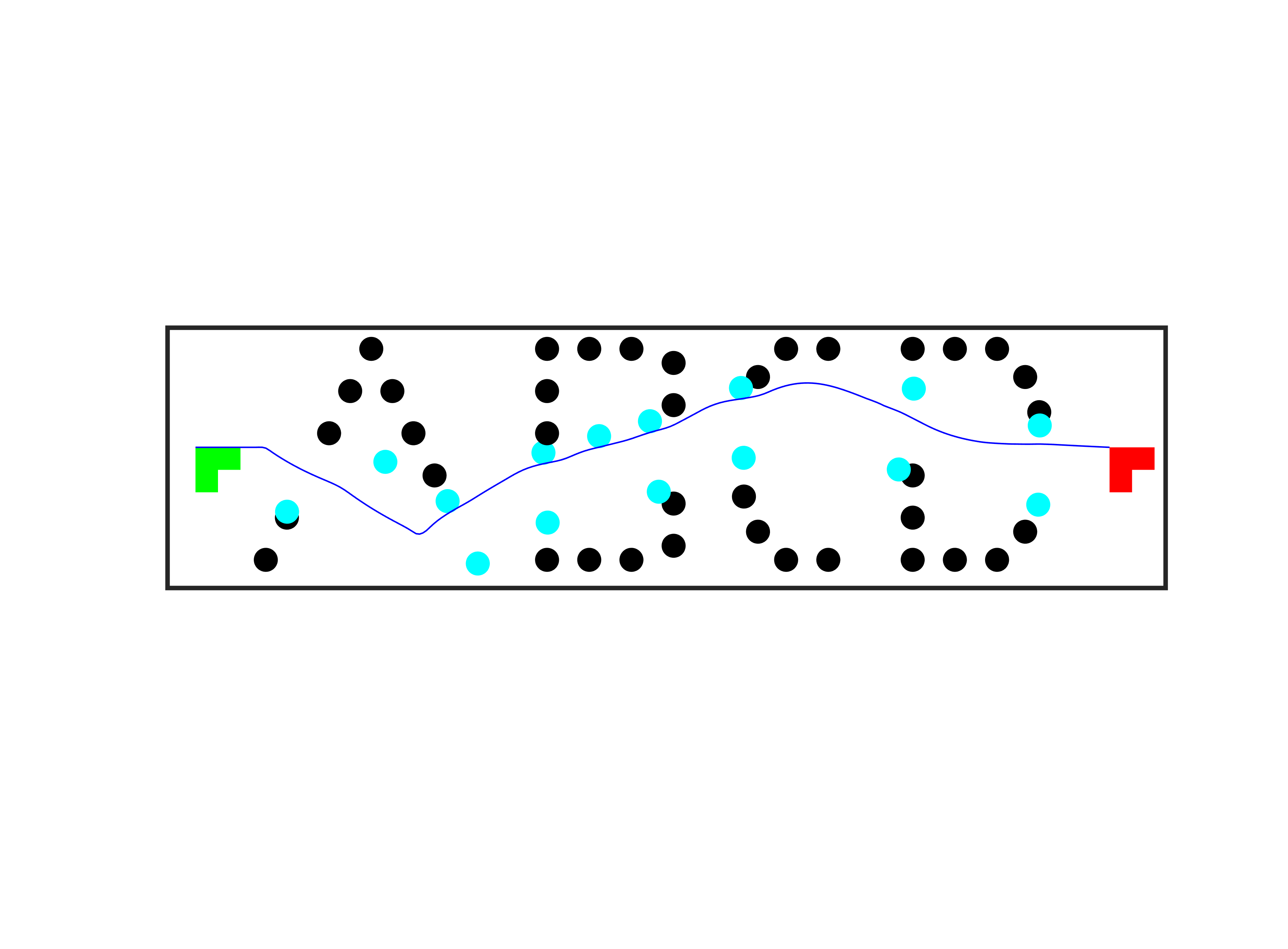}\label{fig:L10}}\hfill
   \caption{ An L-shaped robot in the \textit{ABCD domain} has to move from left to right by displacing circular obstacles. 5 different instances of MCD problem simulated by varying the cost parameters. For each experiment, the simulated trajectory based on the controls obtained in the overlap stage are shown in the first column. The figures in the right column correspond to their respective MCD solutions.}
  \label{fig:L}
\end{figure}
In this Section, we discuss how our approach behaves on a number of examples. 
Unless otherwise mentioned, we use $M_u = 0.1$ and $M_g=10$. 
For each overlapping obstacle, the displacement stage takes an average time of 1 second. 
The performance is evaluated on an Intel{\small\textregistered} Core i7-10510U CPU$@$1.80GHz$\times$8 with 16GB RAM under Ubuntu 18.04 LTS. 

Fig.~\ref{fig:L} shows an L-shaped robot in the \textit{ABCD} domain that must reach the right end by moving circular obstacles (53 in number). 
The robot can move both in the $x$ and $y$ directions and follows the non-linear dynamics model
\begin{equation}
\begin{split}
\dot{x}_x & = ucos(\theta) - vsin(\theta)\\
\dot{x}_y & = usin(\theta) + vcos(\theta)\\
\dot{\theta} & = \omega\\
\end{split}
\label{L-model}
\end{equation}
\noindent where $x = (x_x,x_y,\theta)$ is the robot pose with $u,v$ being the linear velocities and $w$ being the angular velocity. 
We set the control limits $-2.5 \ \textrm{m/s} \leq u,v \leq 2.5 \ \textrm{m/s}$ and $-2.5 \ \textrm{rad/s} \leq w \leq 2.5 \ \textrm{rad/s}$. 
As discussed before, we use FORCESPRO solver to generate optimized nonlinear MPC code~\eqref{eq:cost_fn} and the time taken for the overlap stage is essentially the code generation time. 
We use a time step of $\Delta t = 0.1 s$ and a look-ahead $L=21$ steps. 
We set $M_x=4.07$ and use $M_i=0.3$, $M_i=0.5$ and $M_i=0.7$ to generate 3 different instances of the MCD problem with overall obstacles displacement magnitudes of 20.01 m, 18.63 m and 17.72 m, respectively. 
The simulated trajectories with obstacle overlaps and the corresponding MCD solutions are shown in Fig.~\ref{fig:L1}-\ref{fig:L6}. 
We compare the solutions to the shortest path length problem by ignoring the movable obstacles (no overlap cost is considered). 
This results in a MCD problem as shown in Fig.~\ref{fig:L7}-\ref{fig:L8}, with a displacement magnitude of 21.71 m. 
Thus, in this experiment, if the robot takes the shortest path then the obstacles are to be moved by an additional displacement of about 4 m. 
Fig.~\ref{fig:L9}-\ref{fig:L10} displays different stages of the same problem when the look-ahead is decreased to $L=11$. 
As expected, the solution quality is reduced due to the shorter horizon, giving a displacement magnitude of 19.52 m. 
The different statistics are given in Table~\ref{tab:comp1}. 
\begin{table}[t]
\begin{tabular}{|c|c|c|} 
\hline
Experiment             & Overlap stage (s)  & Displacement magnitude (m) \\ 
\hline
\hline
Shortest path & 3.75          & 21.71 \\ 
$L=11$, $M_x=0.7$        & 6.98          & 19.52\\
$L=21$, $M_x=0.3$        & 13.35         & 20.01\\
$L=21$, $M_x=0.5$        & 13.34          & 18.63\\
$L=21$, $M_x=0.7$        & 13.38          & 17.72\\
\hline
\end{tabular}
\caption{Average computation time for the overlap stage and the displacement magnitudes for different experiments of the MCD problem of L-shaped robot.}
\label{tab:comp1}
\end{table}

\begin{figure}[]
\subfloat[]{\includegraphics[trim=52 120 38 100,clip,scale=0.5]{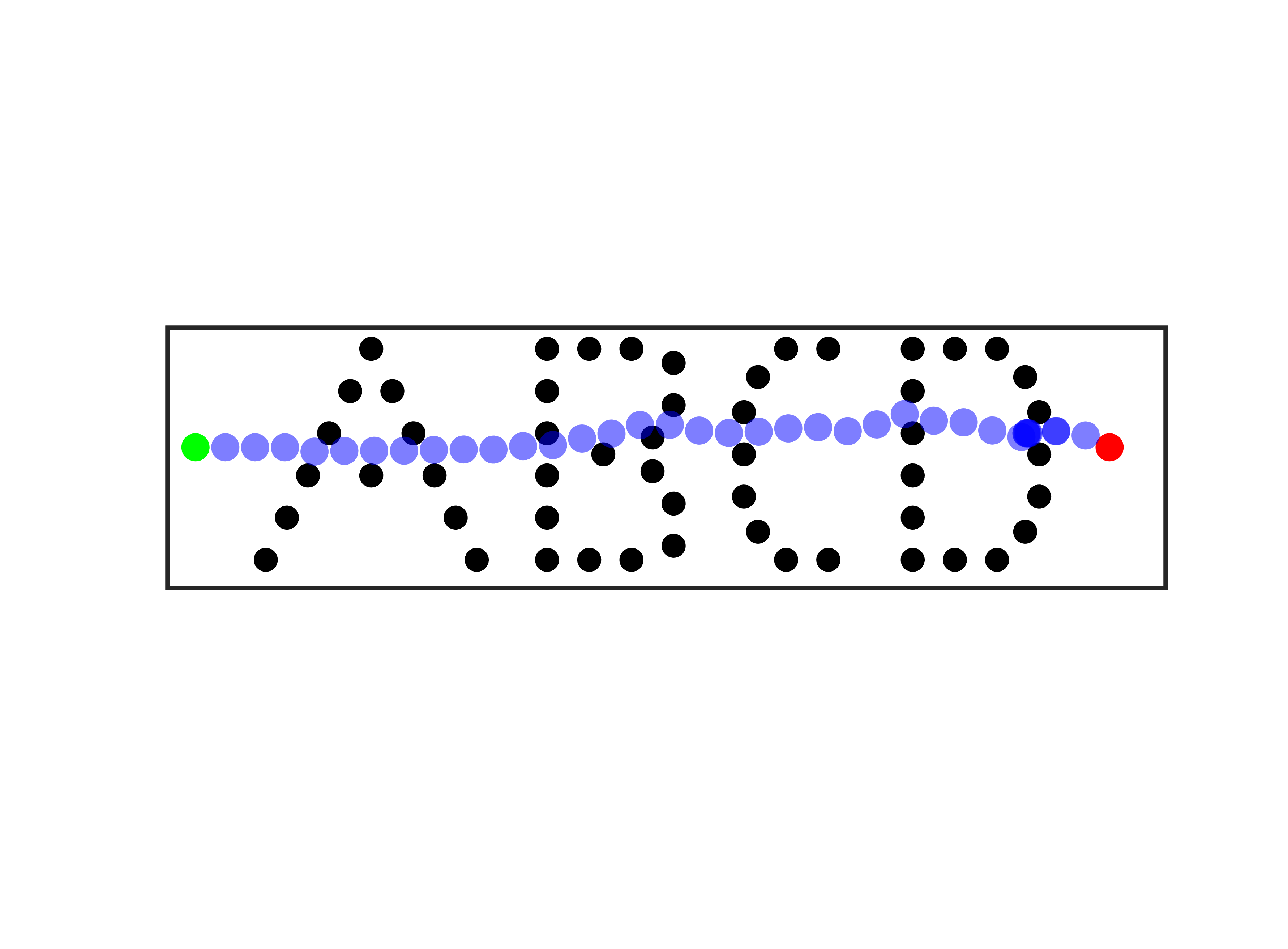}\label{fig:C1}}\hfill
\subfloat[]{\includegraphics[trim=52 120 38 100,clip,scale=0.5]{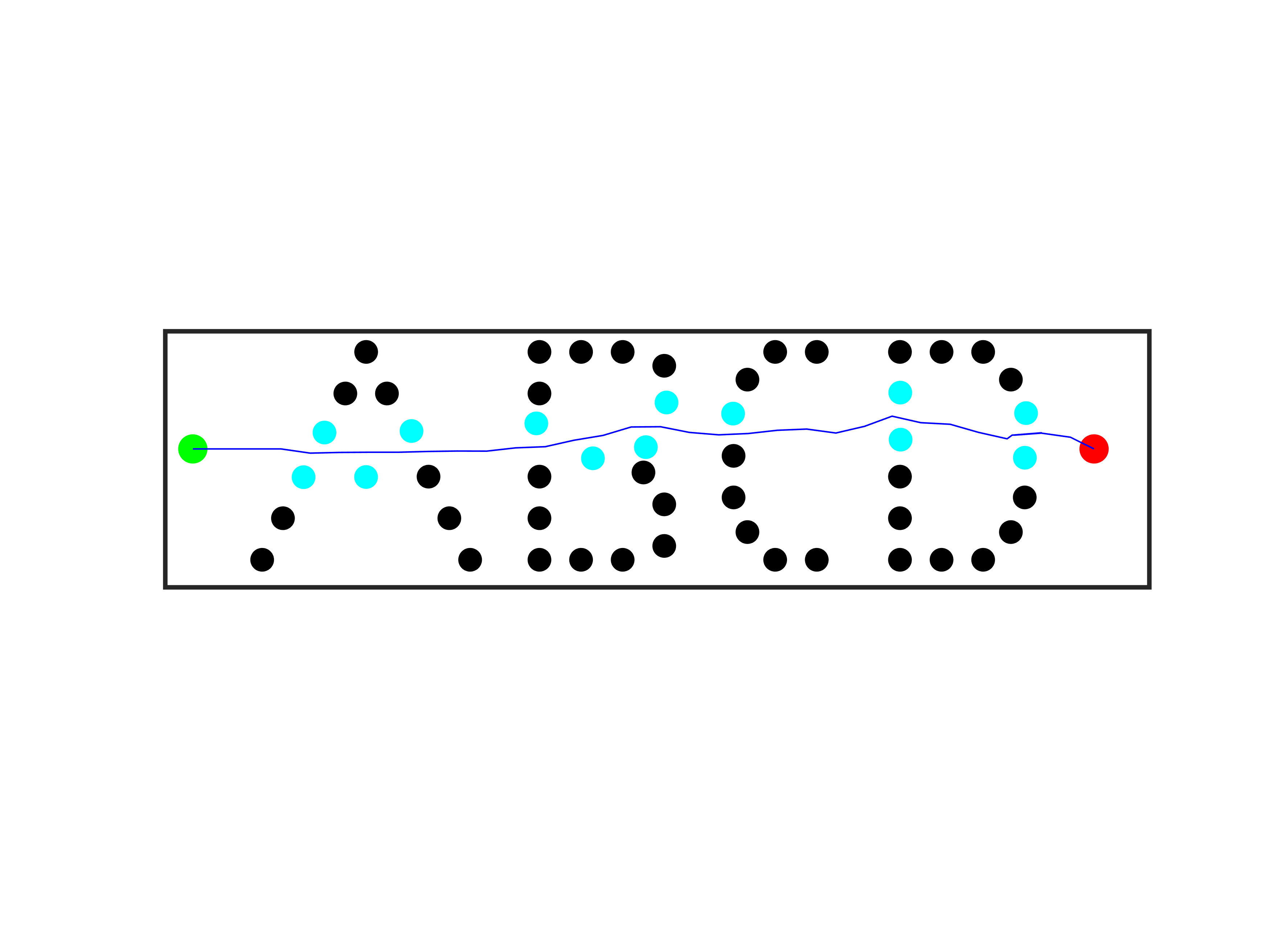}\label{fig:C2}}\hfill
\subfloat[]{\includegraphics[trim=52 120 38 100,clip,scale=0.5]{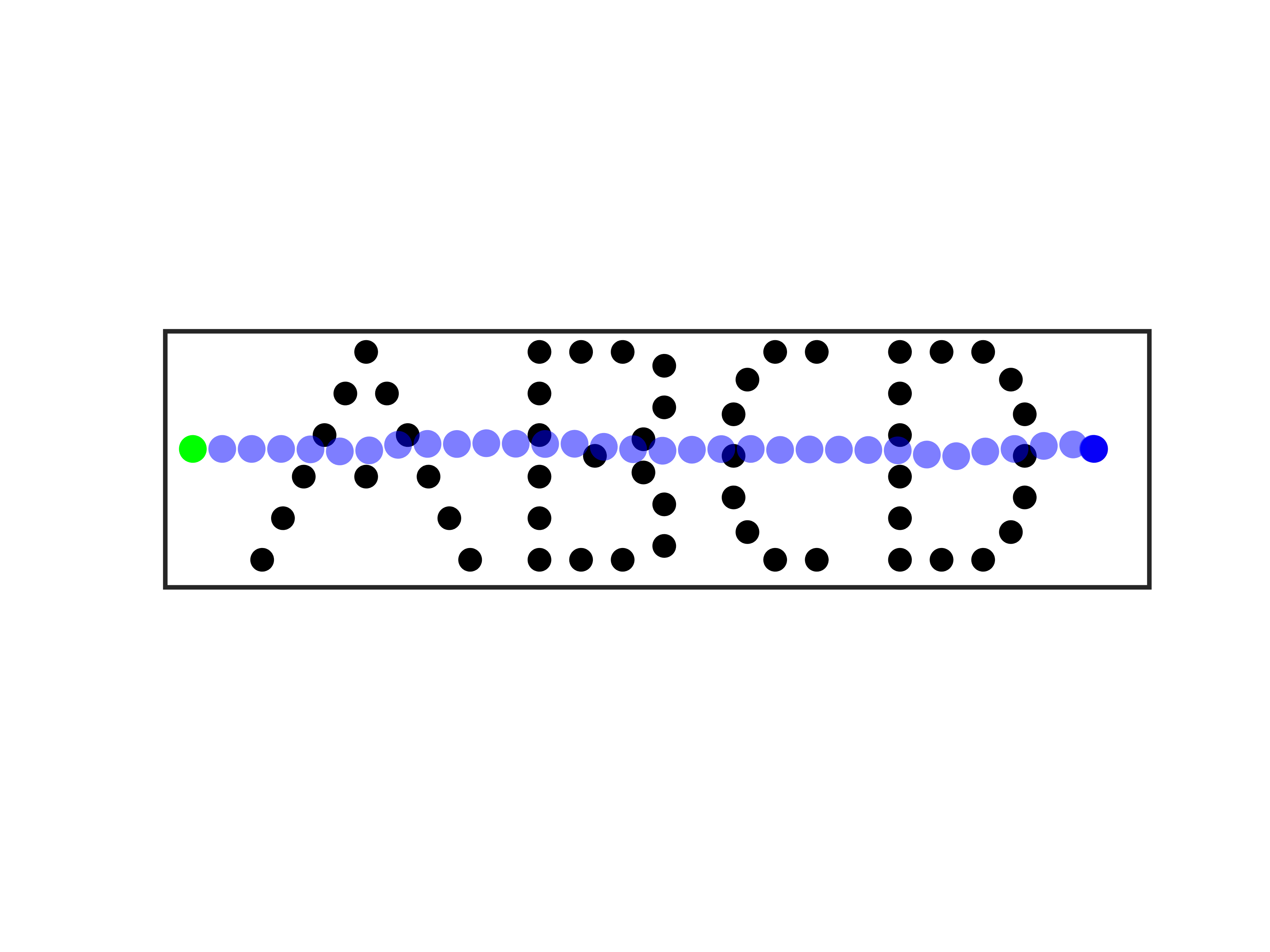}\label{fig:C3}}\hfill
\subfloat[]{\includegraphics[trim=52 120 38 100,clip,scale=0.5]{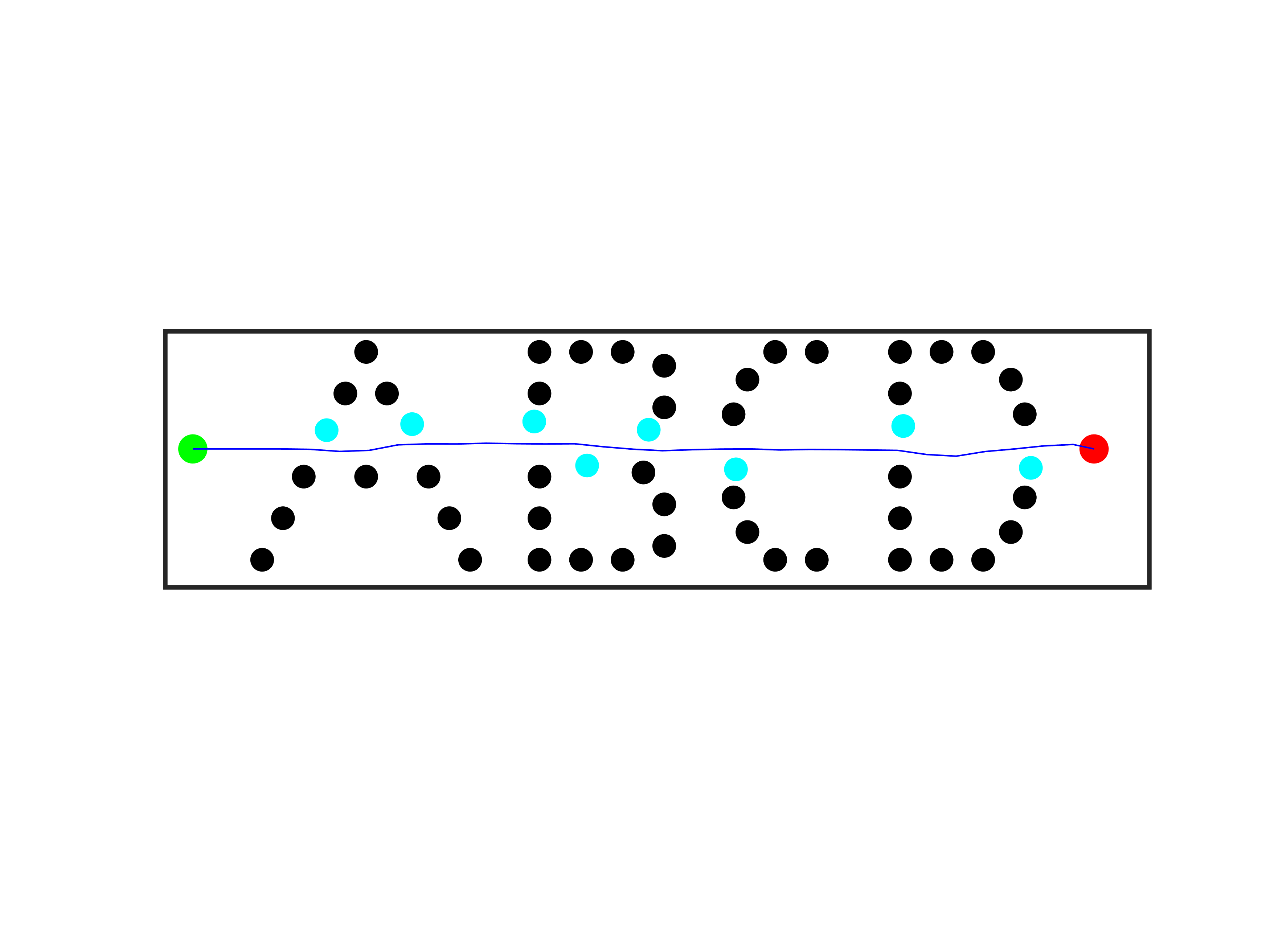}\label{fig:C4}}\hfill
\caption{
A circular robot moving from left to right in the \textit{ABCD domain}. 
(a)-(b) A 13-obstacle solution. 
(c)-(d) An 8-obstacle solution with smaller displacement magnitude.}
\label{fig:C}
\end{figure}

Fig.~\ref{fig:C} demonstrates that a minimum displacement of obstacles does not correspond to the minimum number of displaced obstacles. 
In Fig.~\ref{fig:C1}-\ref{fig:C2} 13 obstacles are displaced with the combined displacement magnitude being 3.45 m. 
An 8-obstacle solution is shown in Fig.~\ref{fig:C3}-\ref{fig:C4}. 
However, compared to the 13-obstacle solution, a greater displacement magnitude of 6.61 m is obtained. 
For this example, we use the following non-linear dynamics for the robot
\begin{equation}
x_{k+1}= \begin{bmatrix}
x_x  \\ 
x_y  \\
\theta \\
\end{bmatrix}_{k+1} = 
\begin{bmatrix}
x_x  \\ 
x_y  \\
\theta \\
\end{bmatrix}_{k} + 
\begin{bmatrix}
D cos(\theta_k + T/2) + C cos(\theta + (T+\pi)/2)  \\ 
D sin(\theta_k + T/2) + C sin(\theta + (T+\pi)/2)  \\
T  \\
\end{bmatrix}
\label{C-model}
\end{equation}
\noindent where $x = (x_x,x_y,\theta)$ is the robot state and $D$, $C$, $T$ are the down-range, cross-range and turn control components, respectively, with a lower bound of [-1.5 m, -1.5 m, -$\pi$ rad] and an upper bound of [1.5 m, 1.5 m, $\pi$ rad]. 
\begin{figure}[]
\subfloat[]{\includegraphics[trim=45 0 0 8,clip,scale=0.2]{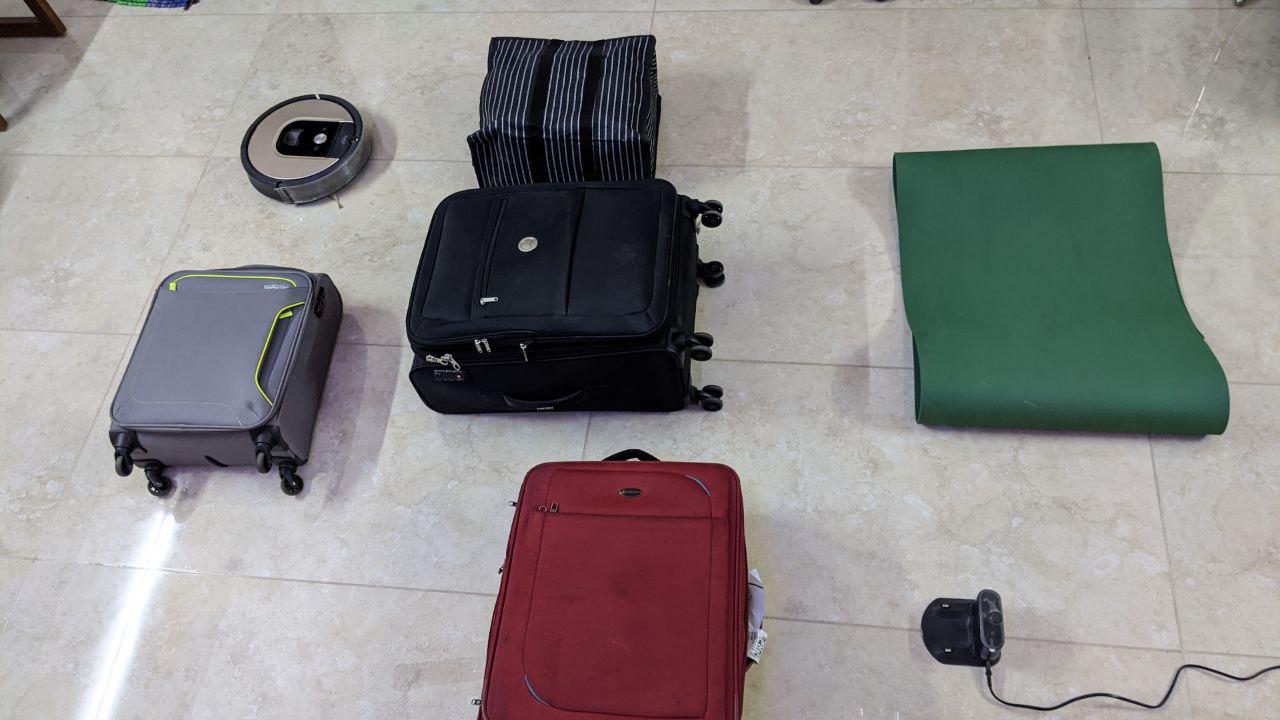}\label{fig:i1}}\hfill
\subfloat[]{\includegraphics[trim=15 40 15 10,clip,scale=0.4]{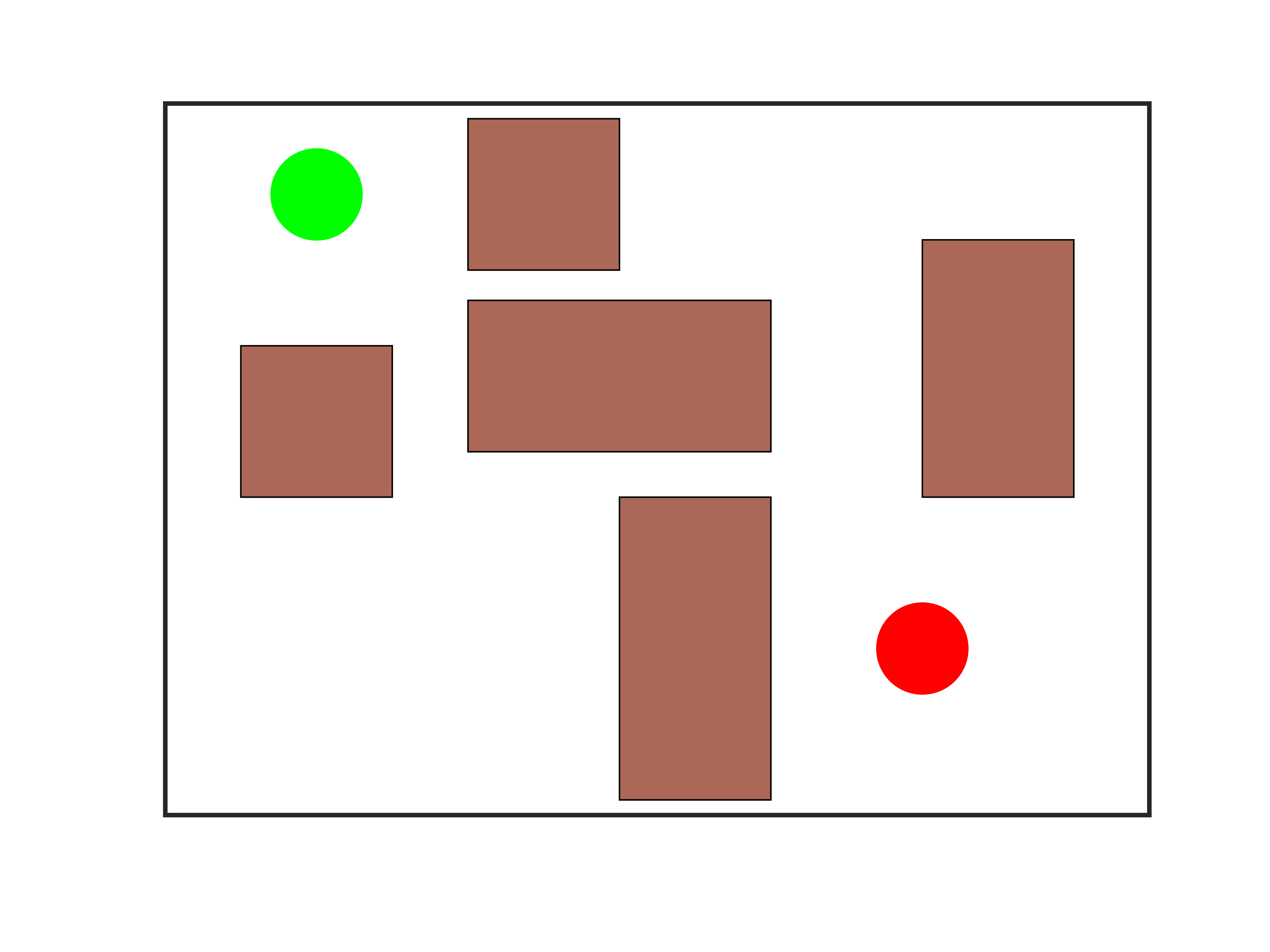}\label{fig:i2}}
\subfloat[]{\includegraphics[trim=15 40 15 10,clip,scale=0.4]{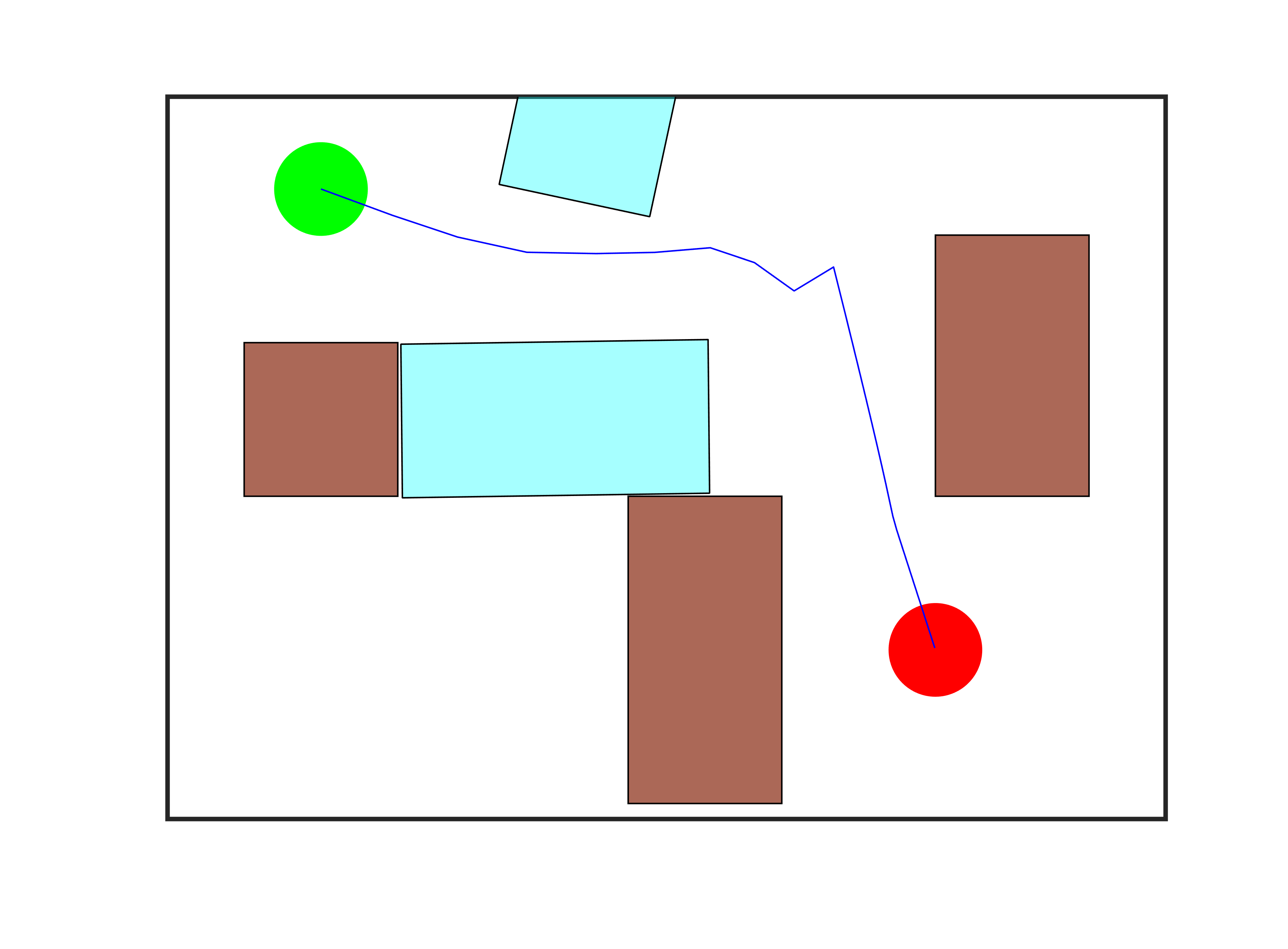}\label{fig:i3}}
\caption{
(a) A representative example of the MCD problem in which a robot has to reach the charging station at the bottom right. 
(b) The map of the environment with objects represented as polygons using planar projections. 
(c) 2-obstacle solution obtained employing the optimization procedure in~(\ref{eq:cpdisp}).}
\label{fig:irobo}
\end{figure}
\begin{figure}[]
\subfloat[]{\includegraphics[trim=35 100 35 20,clip,scale=0.5]{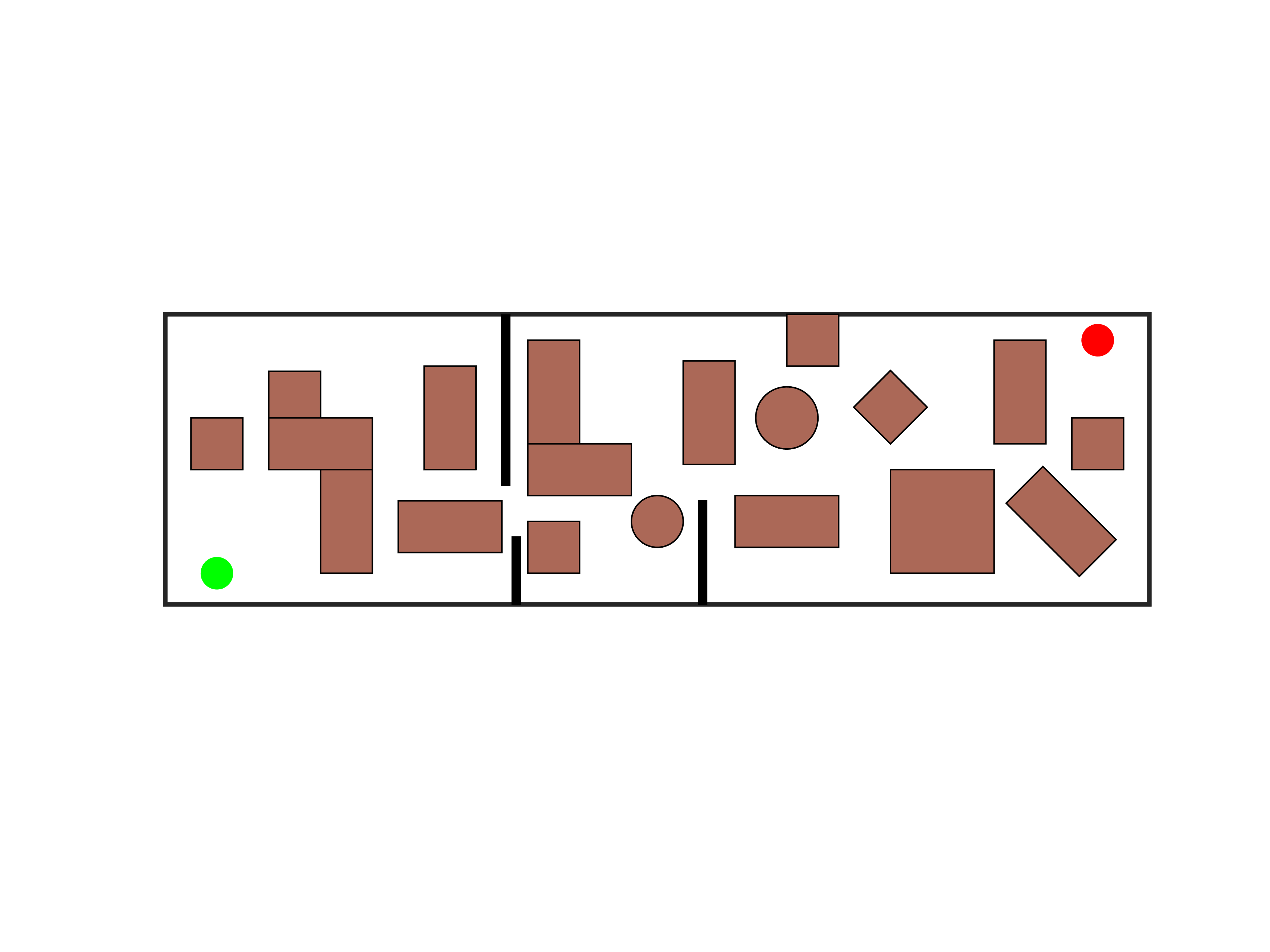}\label{fig:w1}}
\subfloat[]{\includegraphics[trim=35 100 35 20,clip,scale=0.5]{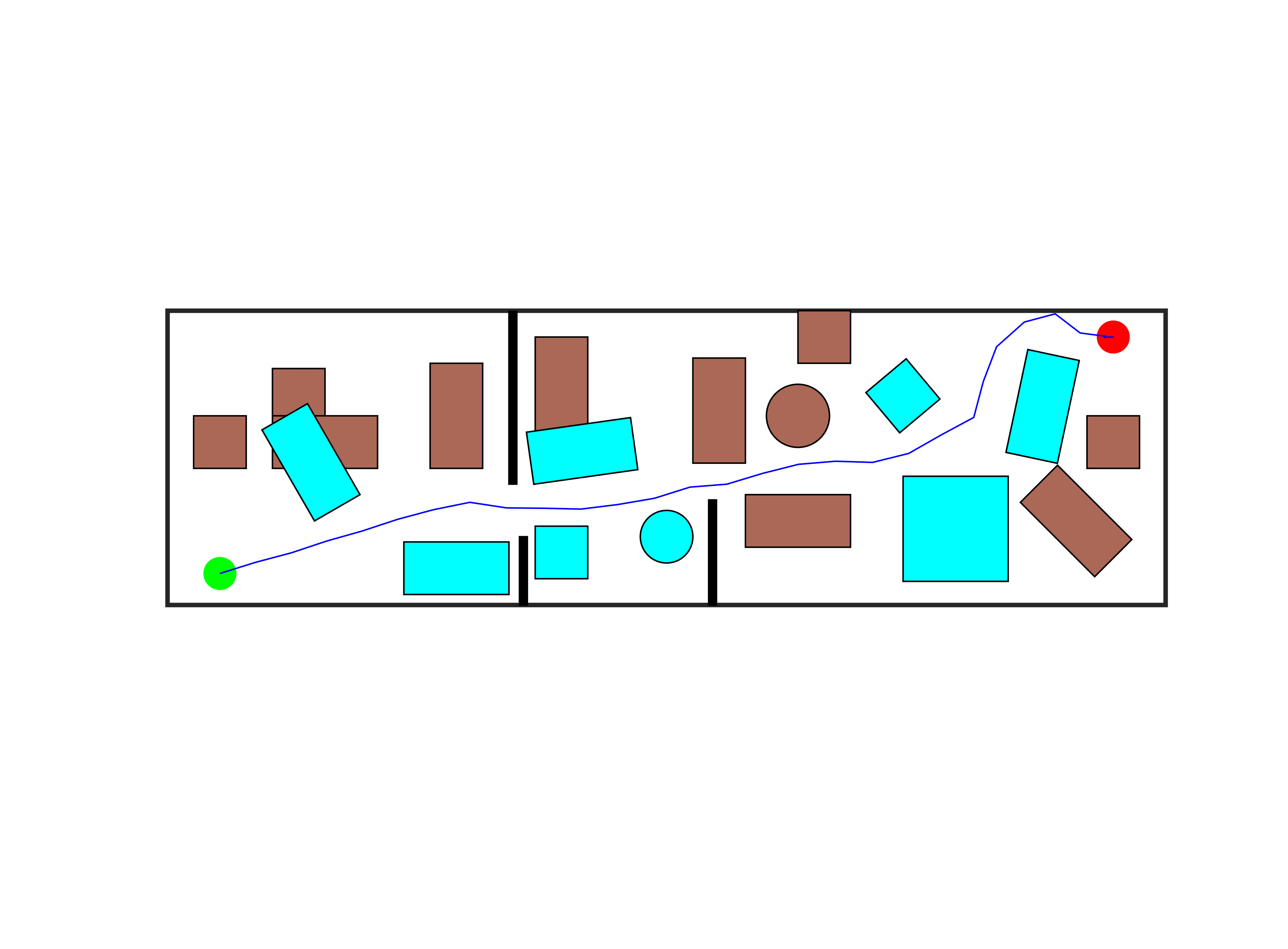}\label{fig:w2}}
\caption{
(a) A representative example of the MCD domain with 19 obstacles. 
(b) An 8-obstacle solution is computed.}
\label{fig:world}
\end{figure}

A sample MCD domain is shown in Fig.~\ref{fig:i1} with an iRobot Roomba robot vacuum cleaner (top left). 
The robot has to reach the charging station (bottom right) by moving obstacles. 
As noted before, in this work we do not focus on how the obstacles are displaced and assume that the computed displacements can be achieved irrespective of the type of robot-obstacle or any required external interaction. 
The MCD problem is solved assuming the robot dynamics in~\ref{C-model}. 
Fig.~\ref{fig:i2} displays a simulated representation of the domain and a 2-obstacle solution with minimum displacement is shown in Fig.~\ref{fig:i3}. 
An example of a larger scale with 19 obstacles is shown in Fig.~\ref{fig:world}. 
While the overlap stage takes 10.2 seconds for Fig.~\ref{fig:irobo}, the computation time for Fig.~\ref{fig:world} is 13.2 seconds. 

\begin{figure}[]
\subfloat[]{\includegraphics[trim=52 110 38 101,clip,scale=0.5]{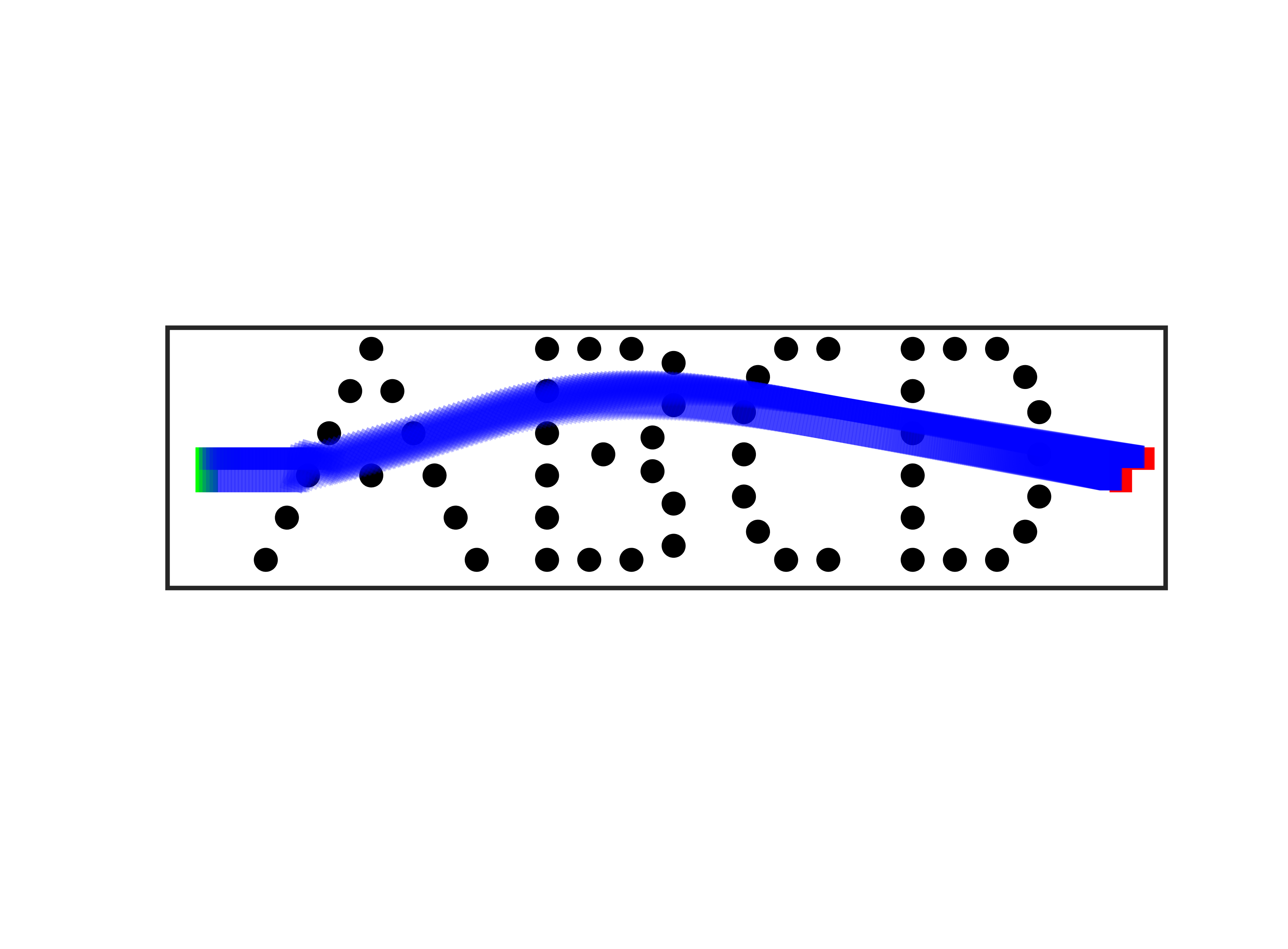}\label{fig:mcr1}}\hfill
\subfloat[]{\includegraphics[trim=52 110 38 101,clip,scale=0.5]{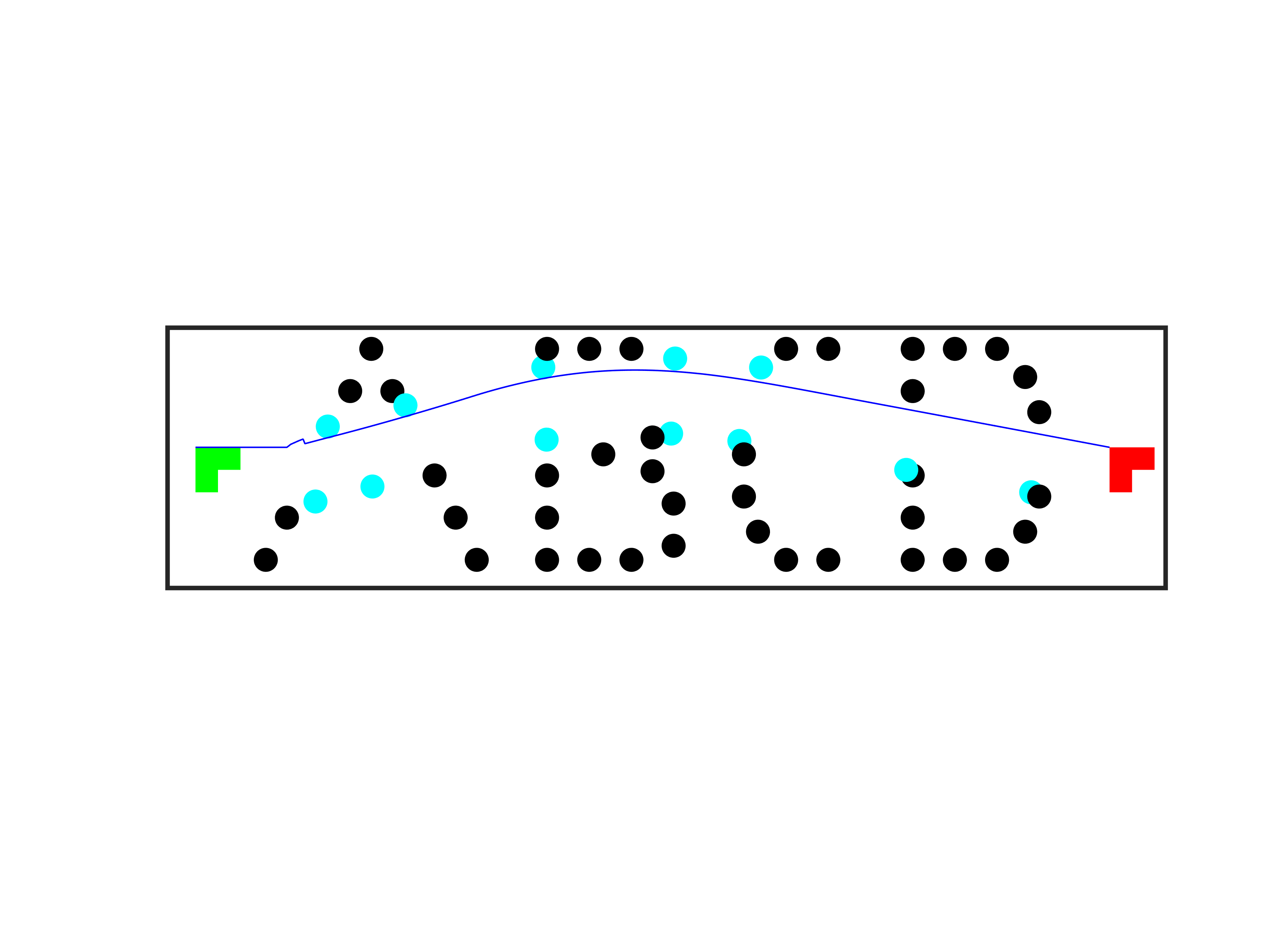}\label{fig:mcr2}}\hfill
\subfloat[]{\includegraphics[trim=52 110 38 101,clip,scale=0.5]{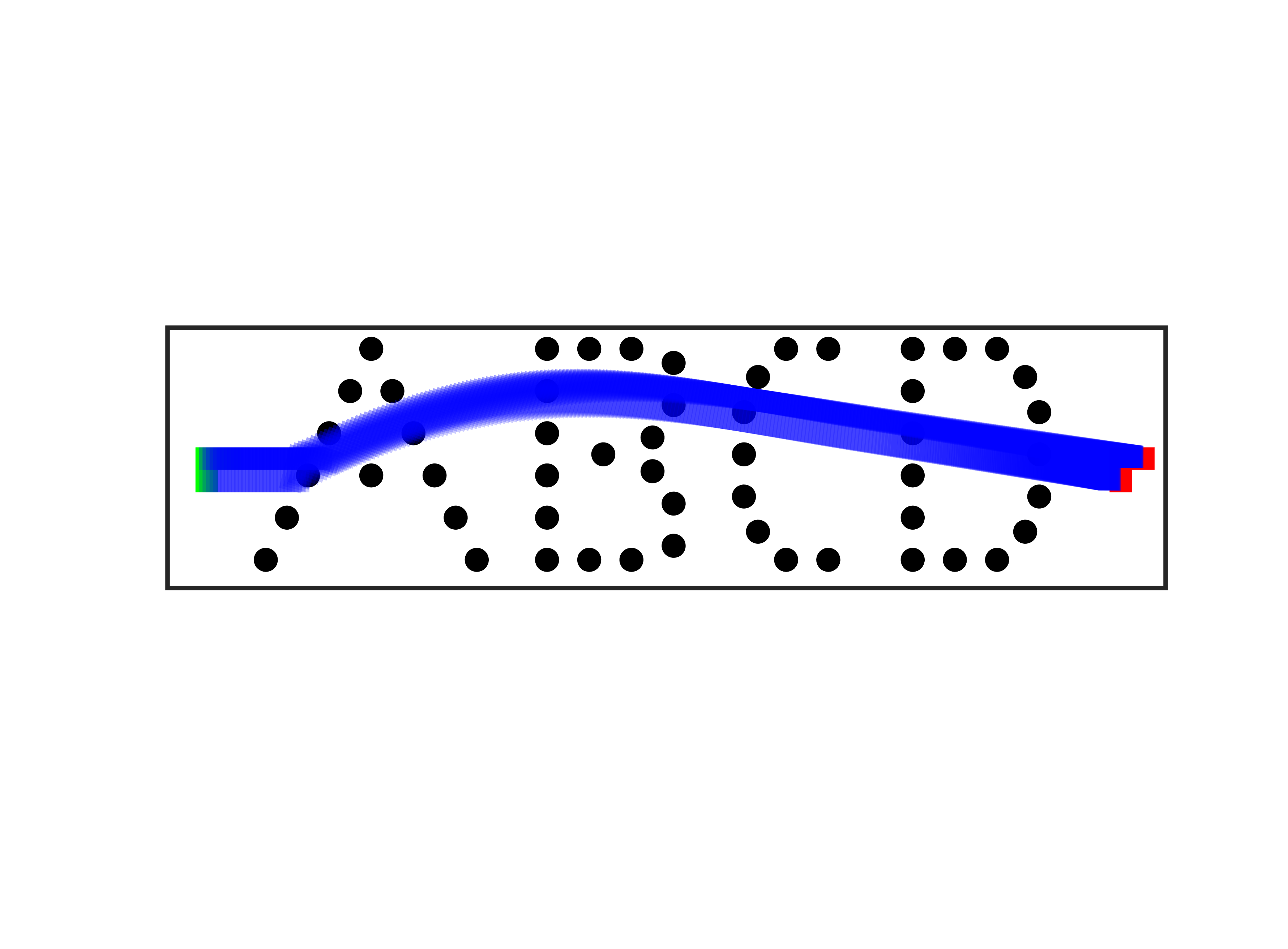}\label{fig:mcr3}}\hfill
\subfloat[]{\includegraphics[trim=52 110 38 101,clip,scale=0.5]{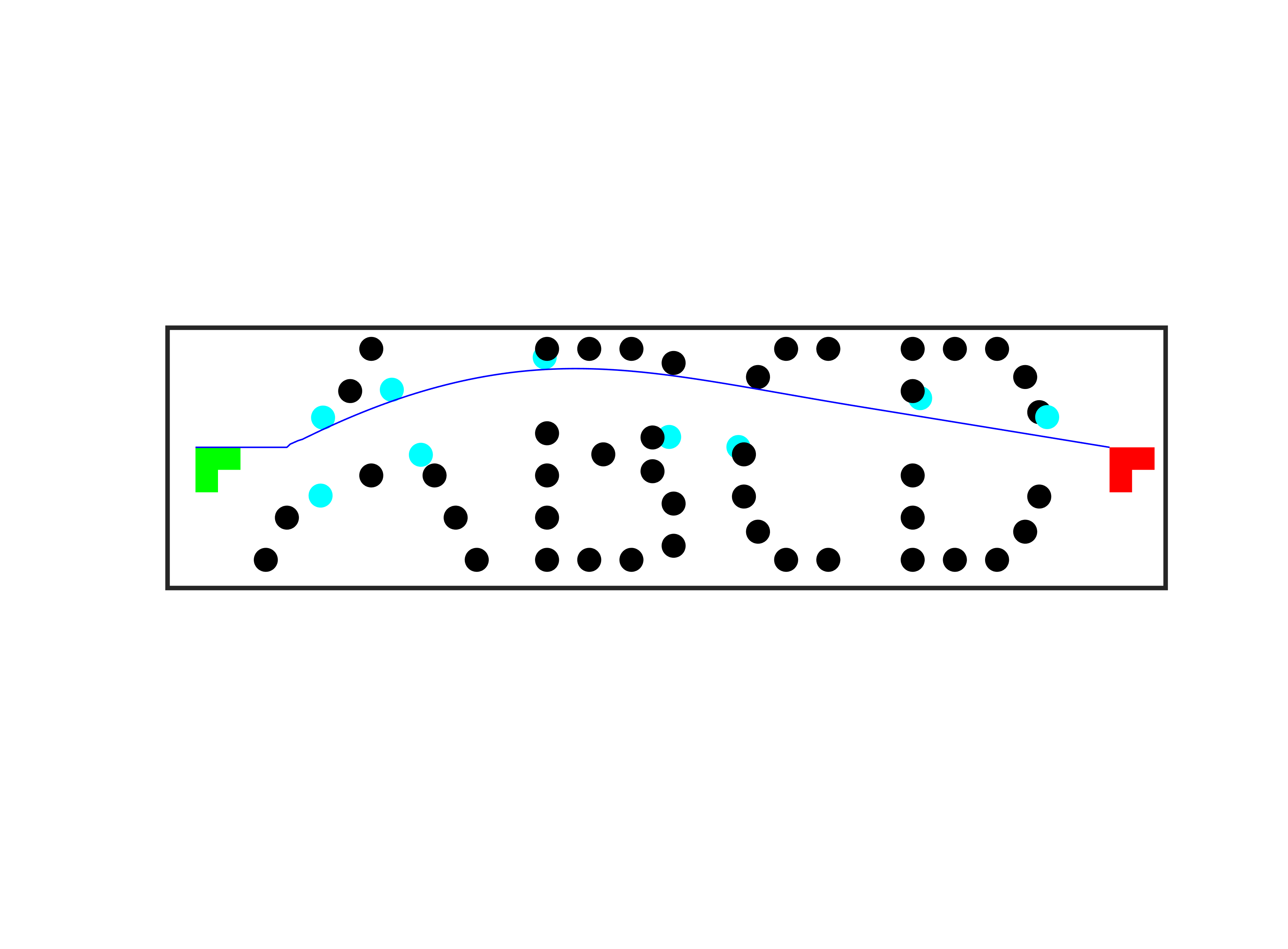}\label{fig:mcr4}}\hfill
\caption{
Illustration of a MCR problem. 
(a)-(b) A 13-obstacle solution. 
(c)-(d) The weight of $h(\mathcal{L}^i)$ is increased to obtain a 9-obstacle solution.}
\label{fig:mcr}
\end{figure}

We now set $h(\mathcal{L}^i) = \eta \frac{\mathcal{L}^i}{\mathcal{L}^i + \epsilon}$, $\epsilon \ll 1$ (to avoid division by zero when $\mathcal{L}^i =0$), to formulate an instance of the MCR problem. 
Initially, all obstacles are assigned a constant value of $\eta = 100$. 
Once an obstacle overlaps with the robot, its $\eta$ is assigned to zero. 
The robot follows the dynamics in~(\ref{L-model}) with the control limits $-2.5 \ \textrm{m/s} \leq u,v \leq 2.5 \ \textrm{m/s}$ and $-2.5 \ \textrm{rad/s} \leq w \leq 2.5 \ \textrm{rad/s}$. 
We use $M_x =0.5$, $M_i=0.5$ and obtain a 12-obstacle solution whose overlapping trajectory and displacement stages are shown in Fig.~\ref{fig:mcr1} and Fig.~\ref{fig:mcr2}, respectively. 
The overlap weight is increased to $M_i=0.7$ and a 9-obstacle solution is obtained, as it can been seen in Fig.~\ref{fig:mcr3}. 
Note that the shortest path solution gives a 13-obstacle solution as already seen in Fig.\ref{fig:L7}. 
Thus, if the shortest path is executed by the robot, then it has to move 13 obstacles as opposed to 9 using our approach. 
Table~\ref{tab:comp2} provides the various statistics.
\begin{table}[t]
\begin{tabular}{|c|c|c|} 
\hline
Experiment             & Overlap stage (s)  & Number of displaced constraints \\ 
\hline
\hline
Shortest path & 3.69          & 13 \\ 
$L=10$, $M_x=0.5$        & 6.87          & 12\\
$L=10$, $M_x=0.9$        & 6.91         & 9\\
\hline
\end{tabular}
\caption{Average computation time for the overlap stage and the number of obstacles to be displaced for the MCR problem.}
\label{tab:comp2}
\end{table}

\section{Discussion}
\label{sec:discussion}
Our approach operates under certain assumptions. 
In the overlap stage, we assume that all obstacles are of equal importance, specifically that each obstacle can be displaced by exerting the same effort. 
This assumption is restrictive in practice (as an example, displacing a heavy table is evidently more difficult than moving a lightweight chair). 
A straightforward way to relax this limitation is to introduce object-specific \textit{difficulty} weights into the overlap computation. 
At present, the weight term ($M_i$) associated with the overlap function in~\eqref{eq:cost_fn} is uniform across all obstacles. 
This can instead be modified to incorporate object-dependent weights reflecting the relative difficulty of displacement.

\begin{figure}[t]
\subfloat[]{\includegraphics[trim=40 103 10 90,clip,scale=0.46]{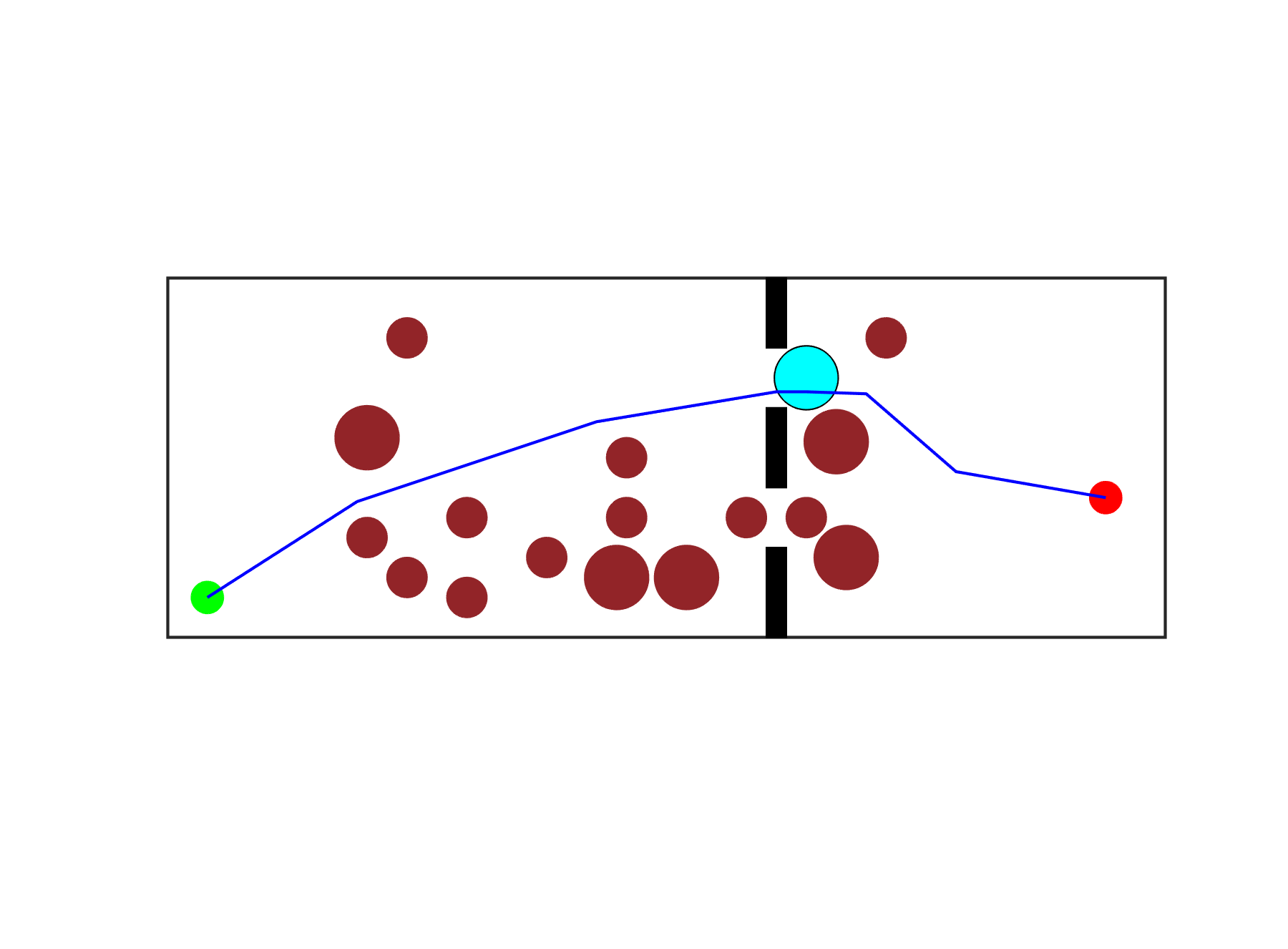}\label{fig:d1}}
\subfloat[]{\includegraphics[trim=40 103 10 90,clip,scale=0.46]{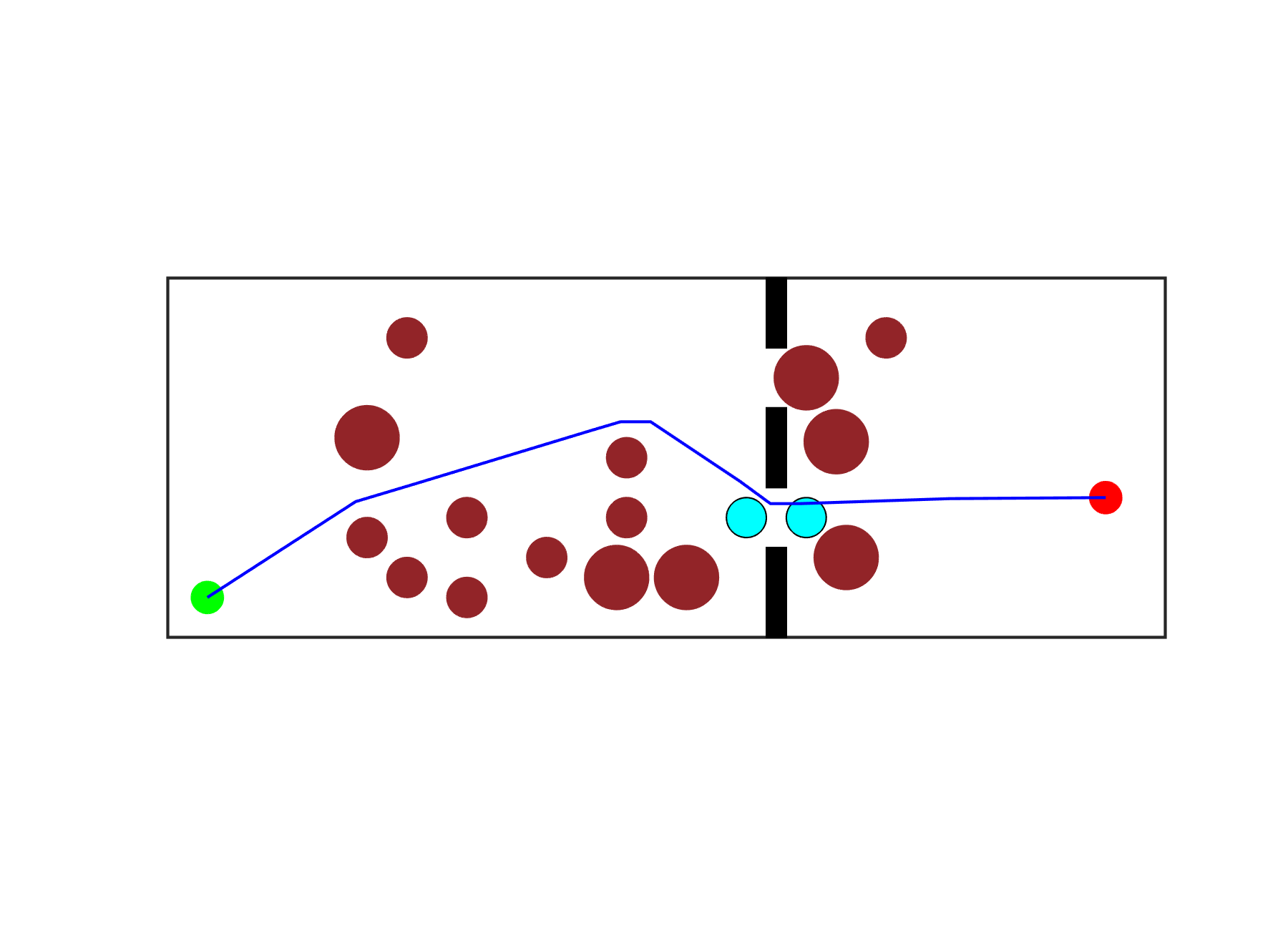}\label{fig:d2}}
\caption{
A robot navigating from the left to the right room in an initially \textit{blocked} environment, where obstacles can be displaced. 
The scene contains 17 obstacles, out of which 6 are larger and heavier. 
(a) The MCR solution when all obstacles are treated equally; a large object must be removed to obtain a feasible path. 
(b) When larger objects are assigned ten times the weight of smaller ones, the MCR solution instead removes two smaller obstacles to find a feasible path.
}
\label{fig:discussion}
\end{figure}

An illustrative example is shown in Fig.~\ref{fig:discussion}.
In the example, adjusting such weights according to each object displacement difficulty results in a different solution to the MCR problem.
A more principled approach would involve incorporating the actual physical work to be done by the robot to move each object into the objective function. 
This would require modeling the manipulation strategy, that is, whether to push or perform a pick-and-place action, along with identifying suitable contact points and feasible trajectories. 
Such considerations, however, are beyond the scope of the present work.

In the displacement stage, the algorithm computes the required displacements of the overlapping obstacles. 
We assume that the computed displacements correspond to actions that can physically carried out by the robot. 
Although this is a reasonable assumption, in practice certain objects may not be movable to the computed locations due to factors such as limited reachability, insufficient payload capacity, or lack of free space to execute manipulation actions.
Furthermore, some objects may be too big or heavy, therefore requiring cooperative manipulation by multiple robots. 
Another practical consideration is the time required to move each object, which could be incorporated into the computation of displacements to prioritize faster-to-move objects or optimize overall task duration. 
Addressing these aspects constitutes a promising direction for future research. 
In cases in which such intricacies are unnecessary, the displacement can alternatively be handled using any state-of-the-art manipulation planning approach~\cite{pan2022TRO, ellis2022IROS}.

\section{Conclusion}
We presented an approach for constraint displacement problems that enables a robot to find a feasible path when constraints or obstacles can be displaced. The approach consists of a two-stage process. It begins by finding a trajectory through the movable obstacles while minimizing an objective function that depends on the specific constraint displacement problem. In the second stage, we calculate locally optimal displacements of the overlapping obstacles. This unified framework can be employed to model various constraint displacement problems as discussed above, using the appropriate objective functions. Section~\ref{sec:results} demonstrated the applicability of our approach to MCD and MCR problems

Currently, we do not consider interacting obstacles, that is, obstacles themselves are not collision-free. We hope to relax this assumption in the future. In future work, we would also like to provide displacement bounds for the constraints so that the total displacements fall within the given bound. We also envision a unified framework that combines the two stages discussed herein. 

%\section*{CRediT authorship contribution statement}
%\textbf{Antony Thomas:} Conceptualization, Methodology, Software, Writing - original draft, Writing - Review \& Editing. \textbf{Fulvio Mastrogiovanni:} Writing - Review \& Editing. \textbf{Marco Baglietto:} Writing - Review \& Editing.
\bibliographystyle{elsarticle-num}
\bibliography{References}
\end{document}